\documentclass[letterpaper,twocolumn,10pt]{article}
\usepackage{usenix}

\usepackage{tikz}
\usepackage{amsmath}

\usepackage{filecontents}

\usepackage{amsmath,amsfonts,bm}

\def\eqref#1{equation~\ref{#1}}

\def\1{\bm{1}}

\DeclareMathAlphabet{\mathsfit}{\encodingdefault}{\sfdefault}{m}{sl}
\SetMathAlphabet{\mathsfit}{bold}{\encodingdefault}{\sfdefault}{bx}{n}

\newcommand{\E}{\mathbb{E}}

\newcommand{\R}{\mathbb{R}}

\DeclareMathOperator{\Tr}{Tr}
\DeclareMathOperator{\divg}{div}

\usepackage{graphicx}
\usepackage{subcaption}
\usepackage{enumitem}
\usepackage{hyperref}
\usepackage{url}
\usepackage{xurl}
\usepackage[conf={restate,no link to proof}]{proof-at-the-end}
\newtheorem{theorem}{Theorem}

\newtheorem{lemma}{Lemma}

\newtheorem{assumption}{Assumption}

\newcommand\blfootnote[1]{%
  \begingroup
  \renewcommand\thefootnote{}\footnote{#1}%
  \addtocounter{footnote}{-1}%
  \endgroup
}

\begin{document}

\date{}

\title{\Large \bf Spectrum-Aware Bounds on Invertibility for Privacy-Enhancing Instance Encoding}

\author{
{\rm Seokjin Hwang$^{*}$}\\
The Pennsylvania State University
\and
{\rm Yuting (Ray) Li$^{*}$}\\
The Pennsylvania State University
\and
{\rm Kiwan Maeng}\\
The Pennsylvania State University
} 

\maketitle

\blfootnote{$^{*}$Equal contribution.}

\begin{abstract}
Instance encoding is a popular empirical technique for privacy enhancement when sharing data to an untrusted server. It transforms sensitive data through an encoding process before sharing, with the hope that the encoding process retains utility but makes it hard to reconstruct the original data. However, most work offers no theoretical guarantee that the encoding process is actually irreversible. A recent work derived a mean-squared error (MSE) bound limiting any adversary's reconstruction accuracy, offering one of the first theoretical results in this domain. This bound, however, has three critical limitations: it is often too loose, only works with randomized encoders (excluding many deterministic encoders practitioners use), and only bounds MSE. We introduce a family of new bounds that (1) are tighter, (2) applicable even to fully deterministic encoders, and (3) can extend beyond MSE to other norm-based similarity metrics, by properly accounting for the encoder's spectral structure. We evaluate our bounds across a range of encoders, datasets, and attacks, showing they hold consistently and improve upon the existing bound.
\end{abstract}

\section{Introduction}
\label{sec:intro}

Machine learning (ML) applications frequently operate on sensitive data during both training and inference. A medical model that predicts a diagnosis from a patient's X-ray image~\cite{x-ray}, for instance, takes protected health information as its input. When training or inference runs on a remote server that users neither trust nor control, users must send their sensitive data to the server---exposing it to \emph{privacy risks}.

\begin{figure}[t]
    \centering
    \includegraphics[width=0.4\textwidth]{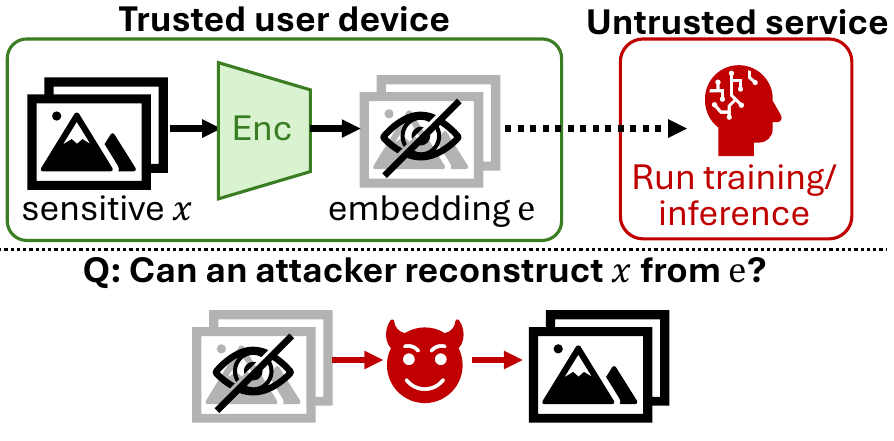}
    \caption{The idea of instance encoding and its invertibility.}
    \label{fig:instance_encoding}
\end{figure}

\emph{Instance encoding}~\cite{instahide_broken} is a popular heuristic approach to improving privacy in such settings. The idea is to process the sensitive data with a privacy-enhancing \emph{encoder} that outputs an \emph{embedding}, and to release the embedding instead of the raw data (Figure~\ref{fig:instance_encoding}).
The hope is that a suitably designed encoder makes it hard to invert the embedding and recover the original data, while still preserving enough information for downstream tasks, such as training or inference.
The same idea appears under a variety of names: data obfuscation~\cite{yala2022syfer, pac_possibility_impossibility}, learnable encryption~\cite{instahide, neuracrypt, dauntless, imaginary_rotate, pac_possibility_impossibility}, split learning~\cite{split_learning, split_learning2}, split inference~\cite{neurosurgeon, splitnets, privynet}, and vertical federated learning (vFL)~\cite{fl_survey, sfl, ressfl} all build on the same idea.

Unfortunately, most existing work in this area is heuristic and lacks a rigorous theoretical foundation: it is unclear what privacy properties, if any, these encoders actually provide. Relying on empirical evaluation alone is unsound, since a scheme that can resist all known attacks and considered safe may later be successfully attacked. In fact, the community has already seen several such failures~\cite{instahide_broken, neuracrypt_broken}. It is therefore important to understand, in a principled way, which privacy properties these encoders do and do not provide, and how the invertibility of an encoder can be quantified.

One notable step in this direction is the work of Maeng et al.~\cite{maeng2023bounding}, who derived a mean-squared error (MSE) bound limiting how precisely an (arbitrarily strong) adversary can reconstruct the original data from its embedding (Section~\ref{sec:fisher2023}).
While this work made important progress towards a theoretical foundation, its bound has several limitations. First, although tight in some cases, the bound becomes very loose in other cases. Second, it applies only to randomized encoders. This renders it inapplicable to many DNN-based encoders proposed by practitioners, which has no randomness~\cite{nopeek, nopeek-infer, ressfl, privynet, disco, yala2022syfer, deepobfuscator, patrol}.
Finally, the bound only constrains the MSE between the original and reconstructed data. MSE, while frequently used as a proxy to data similarity~\cite{maeng2023bounding, guo2022bounding}, is not always a faithful measure of the reconstruction quality: a reconstruction can have high MSE while still leaking sensitive semantic information, or low MSE without meaningful leakage.


In this paper, we provide a set of improved bounds that address these limitations. Our new bounds better account for the spectral geometry of the encoder, yielding tighter bounds in cases where the prior work~\cite{maeng2023bounding} fell short. Our bounds also stay non-trivial even when the encoder induces no randomness, by correctly capturing how the information loss through the encoder's spectral geometry makes reconstruction harder.
Finally, our bounds generalize beyond MSE to other norm-based similarity metrics, such as LPIPS~\cite{lpips} or CLIP score~\cite{clip}.
Our bounds are not only tighter but also match our intuition. The bounds go up (improved privacy) when the encoder discards informative directions, especially those the prior constrains weakly or that the similarity metric (\emph{e.g.}, LPIPS) is most sensitive to.
Below, we summarize our contributions:
\begin{itemize}[noitemsep, leftmargin=*, topsep=0pt]
    \item We improve the bound from Maeng et al.~\cite{maeng2023bounding} (Corollary~\ref{thm:m_bound1}), yielding a much tighter bound for cases where the encoder's randomness is small and privacy mainly stems from the encoder's geometry.
    \item We extend the bound to hold even when the encoder adds no randomness at all (Corollary~\ref{thm:m_bound2}). This adds practicality to the bound, as many encoders used by practitioners do not incorporate any randomness~\cite{nopeek, nopeek-infer, ressfl, privynet, disco, yala2022syfer, deepobfuscator, patrol}.
    \item We extend the bounds to other similarity metrics (Theorem~\ref{thm:psi_bound1} and Corollary~\ref{thm:psi_bound2}), such as LPIPS. This allows us to bound a more semantically meaningful metric than MSE.
    \item We develop an improved method (compared to that of Maeng et al.~\cite{maeng2023bounding}) for estimating the dataset prior, which is required for the bound, using well-trained generative models (\emph{e.g.}, DDPM~\cite{ddpm}). This is not our core contribution but has practical value to the community.
    \item We evaluate our new set of bounds on two image datasets (MNIST~\cite{mnist} and CIFAR-10~\cite{cifar10}), two similarity metrics (MSE and LPIPS~\cite{lpips}), and various encoders, showing that the bounds are effective and tight in many cases. We also show that the bound can serve as a reasonable proxy for measuring the invertibility of privacy-enhancing encoders.
\end{itemize}

\section{Background: Instance Encoding}

Instance encoding refers to a family of techniques that transform a secret input $x \in \R^d$ into an \emph{embedding} $e \in \R^m$ using a privacy-enhancing \emph{encoder} $\mathrm{Enc}: \R^d \rightarrow \R^m$, and share $e$ with untrusted parties instead of $x$ (Figure~\ref{fig:instance_encoding}).
The goal is to design an encoder such that training and inference can still be performed using the embedding $e$, while the original data $x$ cannot be reconstructed from $e$.
While this idea has been popular in the community~\cite{instahide, neuracrypt, cloak, shredder, sfl, ressfl, imaginary_rotate, split_learning, nopeek-infer, he2020transnet, yala2022syfer, liu2020datamix, du2023dp, wang2018not}, most prior work relies on empirical arguments rather than theoretical ones to claim that their encoders are non-invertible, \emph{e.g.}, by testing against a few known attacks.

\textbf{Types of encoders.} Numerous different proposals exist for how to design such a privacy-enhancing encoder. The vast majority of proposals in this area use a deep neural network (DNN) as the encoder~\cite{privynet, neuracrypt, dauntless, ressfl, split_learning, split_learning2, nopeek, nopeek-infer, sfl}, with variations.
Most use a DNN trained with various techniques to (hopefully) resist input reconstruction attacks~\cite{nopeek, nopeek-infer}, while others use DNNs with random weights~\cite{dauntless, neuracrypt, yala2022syfer}.
Some add noise to the encoder to introduce randomness~\cite{noise1, noise2, shredder, cloak, ani, maeng2023bounding, dpfe, du2023dp, wang2018not}, while others use a deterministic DNN~\cite{nopeek, ressfl, privynet, disco, yala2022syfer}.
This paper mostly focuses on such DNN-based encoders, although our bounds are applicable to a broader class of encoders (Section~\ref{sec:m_bound}).

\textbf{Theoretical arguments for instance encoding.}
The field was largely devoid of theoretical analysis~\cite{maeng2023bounding}, but a few recent works have made progress.
Carlini et al.~\cite{instahide_broken} proved a negative result: instance encoding cannot simultaneously achieve strong data indistinguishability and utility~\cite{instahide_broken}.
Thus, one must aim for a weaker guarantee, \emph{e.g.}, non-invertibility, which prohibits near-exact reconstruction of the original data.
This is also why simply applying the popular differential privacy (DP) framework to instance encoding~\cite{du2023dp, wang2018not} and claiming safety is insufficient: DP guarantees data indistinguishability, but any meaningfully strong DP guarantee is incompatible with utility in instance encoding~\cite{instahide_broken}.

Maeng et al.~\cite{maeng2023bounding} were among the first to derive a mean-squared error (MSE) bound that limits the adversary's reconstruction accuracy. 
Our work mainly compares against this prior work, so we discuss it in more detail in Section~\ref{sec:fisher2023}.
There is other relevant work that is largely orthogonal to ours.
One line of work~\cite{liyue18, liyue19, language_mdp, singh2023posthoc, split_and_denoise} instead adopts \emph{metric} DP, guaranteeing that changes within some $\ell_p$ ball of the input cannot be distinguished by looking at the embedding. These approaches have two limitations: (1) in many cases, changes in the $\ell_p$ do not have an interpretable meaning, especially when defined in some latent space~\cite{dp_shield, singh2023posthoc, split_and_denoise}, and their privacy becomes highly empirical; and (2) they cannot easily be applied to DNN-based encoders, since it requires computing the sensitivity (Lipschitz constant) of the DNN encoder, which is known to be NP-hard~\cite{weng2018towards}.
Another line of work~\cite{pac_privacy, pac_possibility_impossibility} proposed a framework called PAC privacy and showed that a bound more general than that of Maeng et al.~\cite{maeng2023bounding} (and ours) is \emph{possible}; however, their possibility results are purely theoretical, and their bound has never been evaluated against realistic attacks.
We do not directly compare with these alternate directions.

\section{Bounding the Encoder's Invertibility Through its Spectral Geometry}
\label{sec:m_bound}




\paragraph{Threat model.}
We assume the problem setup shown in Figure~\ref{fig:instance_encoding}.
Our goal is to estimate how accurately an attacker can reconstruct the input $x \sim \pi$ by observing $e$, through an attack $\hat{x}=\mathrm{Att}(e)$. We assume a strong attacker who knows the encoder $\mathrm{Enc}$ and the data prior $\pi$, and who can leverage this knowledge arbitrarily to mount the best possible attack. This threat model matches that of Maeng et al.~\cite{maeng2023bounding} and is much stronger than the adversaries commonly assumed in other work, which assume that the adversary does not know $\pi$~\cite{guo2022bounding} and/or $\mathrm{Enc}$~\cite{neuracrypt, dauntless, imaginary_rotate, privynet, yala2022syfer}.

\subsection{Bound from Prior Work (Maeng et al.~\cite{maeng2023bounding})}
\label{sec:fisher2023}

Maeng et al.~\cite{maeng2023bounding} proposed one of the first reconstruction bounds in this setup.
Their bound requires the encoder to be differentiable and randomized.
To accommodate DNN-based encoders popular among practitioners~\cite{nopeek, nopeek-infer, ressfl, privynet, disco, yala2022syfer, deepobfuscator, patrol}---which are generally neither differentiable nor randomized---they made two changes to commonly used DNNs: (1) non-differentiable operators (\emph{e.g.}, ReLU or MaxPool) were replaced with differentiable approximations (\emph{e.g.}, GELU or AvgPool), and (2) Gaussian noise was added to the encoder output.
That is, if $\mathrm{Enc}_D(x)$ is the deterministic encoder DNN (with non-differentiable components replaced), the encoder they studied can be expressed as $\mathrm{Enc}(x)=\mathrm{Enc}_D(x) + \mathcal{N}(0, \sigma^2I_m)$.

For this type of encoder, Maeng et al.~\cite{maeng2023bounding} showed
that the mean-squared error (MSE) between the original input $x$ and the
reconstruction $\hat{x}$ obtained by any adversary is lower-bounded in
terms of the encoder's Fisher information matrix (FIM), defined as:
\begin{equation}
    \mathcal{I}_e(x) = \frac{1}{\sigma^2}
      J_{\mathrm{Enc}_{\mathrm{D}}}(x)^\top
      J_{\mathrm{Enc}_{\mathrm{D}}}(x),
\label{eq:fim}
\end{equation}
where $J_{\mathrm{Enc}_{\mathrm{D}}}(x)$ is the Jacobian of
$\mathrm{Enc}_{\mathrm{D}}$ with respect to $x$. Their bound, and all of
ours, rests on the following regularity conditions.
\begin{assumption}\label{asm:standing}
The prior $\pi$ admits a density; the joint density
$p(e, x) = \pi(x)\, p(e \mid x)$ is absolutely continuous in $x$ for
a.e.\ $e$; the score $s_{\pi}(x) = \nabla_{x} \log \pi(x)$ and
$S = \E[s_{\pi}(x)\, s_{\pi}(x)^{\top}]$ exist and are finite; and
$\mathcal{I}_{e}(x)$ exists with
$\mathrm{diag}(\mathcal{I}_{e}(x))^{1/2}$ locally integrable in $x$.
\end{assumption}
While some data may naturally satisfy these assumptions, the distribution of commonly used datasets, \emph{e.g.}, natural images, are not absolutely continuous. Maeng et al.~\cite{maeng2023bounding} therefore worked with \emph{smoothed} images~\cite{randomized_smoothing} instead, where a small amount of Gaussian noise was added to each image. Our evaluation also uses such smoothed images. 
Under
Assumption~\ref{asm:standing}, Maeng et al.~\cite{maeng2023bounding}
showed the following.
\begin{theorem}[Bound from~\cite{maeng2023bounding}]
\label{thm:2023_bound}
Under Assumption~\ref{asm:standing}, suppose further that
$p(e, x) \to 0$ as $\|x\| \to \infty$ for a.e.\ $e$, and that
$\E\big[\Tr(\mathcal{I}_e(x))\big] + \Tr(S)$ is finite and nonzero.
For a $d$-dimensional input $x$, the following holds for any
$\hat{x} = \mathrm{Att}(e)$:
\begin{equation}
\E\big[\|\hat{x} - x\|_2^2 / d\big] \ge
  \frac{d}{\E\big[\Tr(\mathcal{I}_e(x))\big] + \Tr(S)}.
\label{eq:2023_bound}
\end{equation}
\end{theorem}


\paragraph{Limitations of Maeng et al.~\cite{maeng2023bounding}}

The bound from Theorem~\ref{thm:2023_bound} has critical issues that limit its applicability.
First, the bound becomes smaller as less noise $\sigma$ is added, and becomes trivial (\emph{i.e.}, zero) when $\sigma=0$.
This may be tight for encoders that are genuinely easy to invert without much noise, but it is very loose for others; practitioners have shown that many encoders, especially DNN-based ones, are hard to invert even when little or no randomness is added~\cite{nopeek, nopeek-infer, ressfl, privynet, disco, yala2022syfer, deepobfuscator, patrol}.
For these encoders, the bound incorrectly suggests that the encoder is easy to invert (because little or no noise is added), when in fact it is not.

Second, the bound only constrains the MSE of the reconstruction. While MSE between the original and reconstructed inputs has been used as a proxy for reconstruction quality in much prior work~\cite{guo2022bounding, maeng2023bounding, hannun2021measuring}, it is known that MSE is not an accurate measure of semantic similarity.
Other similarity metrics are considered better proxies for semantic similarity---such as LPIPS~\cite{lpips} or CLIP score~\cite{clip} for images, and sentence embedding similarity~\cite{rag} for natural language.
We introduce a set of alternative bounds that address these issues.

\subsection{Bounds with the Spectral Geometry}
\label{sec:m_bound1}

First, we introduce a new bound that is tighter than the bound of Maeng et al.~\cite{maeng2023bounding} (Theorem~\ref{thm:2023_bound}) when the noise added to the encoder is small. Unlike Theorem~\ref{thm:2023_bound}, which relies on a scalar summary of the encoder's spectral geometry (through the term $\Tr(\mathcal{I}_e(x))$), our bound makes better use of the full spectral geometry information.
We derive the bound by extending the van Trees inequality, and hence present it as a Corollary.
Maeng et al.~\cite{maeng2023bounding} also derived their bound from the van Trees inequality~\cite{gill1995applications}, but we use an alternative matrix form~\cite{bouchard2024intrinsic}. For completeness, the form we use is restated in Appendix~\ref{app:vantrees_bouchard2024}, and the proof is given in Appendix~\ref{sec:proofs}.

\begin{theoremEnd}[proof end]{corollary}[Improved MSE bound]
\label{thm:m_bound1}
Under Assumption~\ref{asm:standing}, suppose further that
$x\, p(e, x) \to 0$ as $\|x\| \to \infty$ for a.e.\ $e$, and that
$\E\big[\mathcal{I}_e(x)\big] + S$ is non-singular. Then the following
holds for any attack $\hat{x} = \mathrm{Att}(e)$:
\begin{equation}
    \E\big[\|\hat{x} - x\|_2^2 / d\big] \ge
      \frac{1}{d} \Tr\Big(
        \big( \E\big[\mathcal{I}_e(x)\big] + S \big)^{-1} \Big).
\label{eq:bound1}
\end{equation}
\end{theoremEnd}
Again, using smoothed images (or any other dataset that naturally meets the assumptions) allow us to use the bound.

\begin{proofEnd}
    Replacing $\theta$ with $x$ and $y$ with $e$ from Lemma~\ref{thm:vantrees1} immediately gives the result. Note that for our randomized DNN encoder,
\[
\E_{e|x}\big[\nabla_x \log p(e|x) \nabla_\theta \log p(e|x)^\top\big] = \frac{1}{\sigma^2}J_{\mathrm{Enc_{D}}}(x)^\top J_{\mathrm{Enc_{D}}}(x),
\]
which makes the two definitions $\mathcal{I}_y(\theta)$ from Lemma~\ref{thm:vantrees1} and $\mathcal{I}_e(x)$ from Equation~\ref{eq:fim} to collapse to the same matrix under notational changes (hence, we use the same notation $\mathcal{I}$ for both).
\end{proofEnd}

The bound from Corollary~\ref{thm:m_bound1} looks similar to that from Theorem~\ref{thm:2023_bound} but has one crucial difference that makes it tighter for smaller $\sigma$: Theorem~\ref{thm:2023_bound} takes the \emph{inverse of the trace}, while our bound (Corollary~\ref{thm:m_bound1}) takes the \emph{trace of the inverse}.
To highlight this difference, simply let $S=0$, and let $\lambda_1, \lambda_2, \ldots, \lambda_d$ be the eigenvalues of $\E\big[\mathcal{I}_e(x)\big]$.
Since the trace of a square matrix equals the sum of its eigenvalues, the right-hand side of Theorem~\ref{thm:2023_bound} becomes $\frac{d}{\lambda_1 + \lambda_2 + \cdots + \lambda_d}$, whereas the right-hand side of Corollary~\ref{thm:m_bound1} becomes $\frac{1}{d}\left(\frac{1}{\lambda_1}+\frac{1}{\lambda_2}+\cdots+\frac{1}{\lambda_d}\right)$. The two become identical when all eigenvalues are equal (\emph{i.e.}, when the encoder is isotropic), but otherwise, the latter is always at least as large.

Colloquially, an encoder can be seen as projecting the input into an embedding space, where the eigenvalues of its (expected) FIM describe how strongly information along each direction survives.
An isotropic encoder preserves information equally along every direction. An anisotropic encoder, on the other hand, amplifies certain directions while attenuating or discarding others---the latter intuitively enhances privacy better.
Our bound is tighter for anisotropic encoders because it explicitly accounts for this directional structure (\emph{e.g.}, the encoder's spectral geometry), correctly predicting that some encoders are harder to invert---even without much noise---if they discard enough directional information.
Later, in Section~\ref{sec:eval}, we show that our new bound (Corollary~\ref{thm:m_bound1}) is indeed much tighter than that from prior work (Theorem~\ref{thm:2023_bound}).

\subsection{Bound in the Case of Zero Noise Addition}
\label{sec:m_bound2}

Our new bound (Corollary~\ref{thm:m_bound1}) still collapses to zero when there is no noise at all ($\sigma=0$). This makes the bound still inapplicable to many privacy-enhancing encoders that do not add any noise.
We therefore extend Corollary~\ref{thm:m_bound1} to cover such noise-free encoders.
For notational simplicity, we define $G(x)=J_{\mathrm{Enc_{D}}}(x)^\top J_{\mathrm{Enc_{D}}}(x)$ (\emph{i.e.}, $\mathcal{I}_e(x) = \frac{1}{\sigma^2}G(x)$).
The proof is given in Appendix~\ref{sec:proofs}.

\begin{theoremEnd}[proof end]{corollary}[Noise-free MSE bound]\label{thm:m_bound2}
Let $r = \mathrm{rank}(\E[G])$, and let
$U \in \R^{d \times (d-r)}$ be an orthonormal basis for
$\mathrm{null}(\E[G])$. Then, for the noise-free deterministic encoder $e=\mathrm{Enc}_{D}(x)$,
\begin{align}
\mathbb{E}\big[\|\hat{x} - x\|_2^2/d\big]
\ge
\frac{1}{d}\Tr\Big((U^\top S U)^{-1}\Big).
\label{eq:bound2}
\end{align}
When $\E[G]$ is full rank ($r=d$), the bound collapses to zero.
\end{theoremEnd}
\begin{proofEnd}
Since $U$ spans $\mathrm{null}(\E[G])$,
\begin{align*}
0
&= \Tr\big(U^\top \E[G]U\big) \\
&= \E\Big[\Tr\big(U^\top J_{\mathrm{Enc}_{D}}(x)^\top
J_{\mathrm{Enc}_{D}}(x)U\big)\Big] \\
&= \E\big[\|J_{\mathrm{Enc}_{D}}(x)U\|_F^2\big].
\end{align*}
Therefore, $J_{\mathrm{Enc}_{D}}(x)U=0$ almost surely under $\pi$.
Since $\pi$ has full support and $J_{\mathrm{Enc}_{D}}$ is continuous, this implies
$J_{\mathrm{Enc}_{D}}(x)U=0$ for every $x$.
Hence, for any $a\in\R^{d-r}$, from the fundamental theorem of calculus,
\begin{align*}
\mathrm{Enc}_{D}(x+Ua)-\mathrm{Enc}_{D}(x)
&=
\int_0^1 J_{\mathrm{Enc}_{D}}(x+tUa)Ua\,dt \\
&= 0.
\end{align*}
In other words, the noise-free encoder is invariant along the directions spanned by $U$.

Let $R$ be an orthonormal basis for $\mathrm{range}(\E[G])$, so that
$[R\;\;U]$ is an orthonormal basis, and write $v=R^\top x$ and $z=U^\top x$.
Because $\mathrm{Enc}_{D}(x)$ is invariant to $z$ (\emph{i.e.}, the encoder does not retain information along the $z$ direction), for any attack
$\hat{x}=\mathrm{Att}(e)$, the projected reconstruction $U^\top\hat{x}$ can
depend on $v$ but not on $z$.
For $s_\pi(x)=\nabla_x\log\pi(x)$, we now show that
\begin{align*}
\E\Big[
U^\top(\hat{x}-x)
\big(U^\top s_\pi(x)\big)^\top
\Big]
=
I_{d-r}.
\end{align*}
Since $[R\;\;U]$ is orthonormal, we have $x=Rv+Uz$, and this change of
variables has unit Jacobian.
Moreover,
\begin{align*}
U^\top s_\pi(x)
&=
U^\top\nabla_x\log\pi(x) \\
&=
\nabla_z\log\pi(v,z).
\end{align*}
Thus, for each $j\in\{1,\ldots,d-r\}$,
\begin{align*}
\big(U^\top s_\pi(x)\big)_j
&=
\frac{\partial}{\partial z_j}\log\pi(v,z) \\
&=
\frac{1}{\pi(v,z)}
\frac{\partial \pi(v,z)}{\partial z_j}.
\end{align*}
Now, consider the $(i,j)$-th entry of
\begin{align*}
\E\Big[
U^\top(\hat{x}-x)
\big(U^\top s_\pi(x)\big)^\top
\Big].
\end{align*}
Since $U^\top x=z$, this entry is
\begin{align*}
&\E\Big[
\big((U^\top\hat{x})_i-z_i\big)
\big(U^\top s_\pi(x)\big)_j
\Big] \\
&=
\int
\big((U^\top\hat{x})_i-z_i\big)
\frac{1}{\pi(v,z)}
\frac{\partial \pi(v,z)}{\partial z_j}
\pi(v,z)
\,dv\,dz \\
&=
\int
\big((U^\top\hat{x})_i-z_i\big)
\frac{\partial \pi(v,z)}{\partial z_j}
\,dv\,dz.
\end{align*}

We now integrate by parts with respect to $z_j$, while treating $v$ and
the remaining coordinates of $z$ as fixed. This gives
\begin{align*}
&\int
\big((U^\top\hat{x})_i-z_i\big)
\frac{\partial \pi(v,z)}{\partial z_j}
\,dv\,dz \\
&=
\int
\left[
\big((U^\top\hat{x})_i-z_i\big)
\pi(v,z)
\right]_{z_j=-\infty}^{z_j=\infty}
\,dv\,dz_{-j} \\
&\quad
-
\int
\frac{\partial}{\partial z_j}
\big((U^\top\hat{x})_i-z_i\big)
\pi(v,z)
\,dv\,dz,
\end{align*}
where $dz_{-j}$ denotes integration over all coordinates of $z$ except
$z_j$.
The boundary term vanishes by the tail condition on $\pi$. Therefore,
\begin{align*}
&\E\Big[
\big((U^\top\hat{x})_i-z_i\big)
\big(U^\top s_\pi(x)\big)_j
\Big] \\
&=
-
\int
\frac{\partial}{\partial z_j}
\big((U^\top\hat{x})_i-z_i\big)
\pi(v,z)
\,dv\,dz.
\end{align*}

From the encoder invariance established above,
$\mathrm{Enc}_{D}(Rv+Uz)$
does not depend on $z$, and thus $\hat{x}=\mathrm{Att}(\mathrm{Enc}_{D}(x))$ and $U^\top\hat{x}$ also cannot depend on $z$. As a result,
\begin{align*}
\frac{\partial}{\partial z_j}(U^\top\hat{x})_i
=
0.
\end{align*}
On the other hand,
\begin{align*}
\frac{\partial z_i}{\partial z_j}
=
\delta_{ij}=
\begin{cases}
1, & i=j,\\
0, & i\neq j.
\end{cases}
\end{align*}
Hence,
\begin{align*}
\frac{\partial}{\partial z_j}
\big((U^\top\hat{x})_i-z_i\big)
=
-\delta_{ij}.
\end{align*}
Substituting this into the previous expression,
\begin{align*}
&\E\Big[
\big((U^\top\hat{x})_i-z_i\big)
\big(U^\top s_\pi(x)\big)_j
\Big] \\
&=
-\int
(-\delta_{ij})
\pi(v,z)
\,dv\,dz \\
&=
\delta_{ij},
\end{align*}
since $\pi$ integrates to one.
Therefore, every $(i,j)$-th entry equals $\delta_{ij}$, and hence
\begin{align*}
\E\Big[
U^\top(\hat{x}-x)
\big(U^\top s_\pi(x)\big)^\top
\Big]
=
I_{d-r}.
\end{align*}

Now let $a=U^\top(\hat{x}-x)$ and $b=U^\top s_\pi(x)$. The identity above
gives $\E[ab^\top]=I_{d-r}$, while
\begin{align*}
\E[bb^\top]
&=
U^\top\E[s_\pi(x)s_\pi(x)^\top]U \\
&=
U^\top S U.
\end{align*}
By the matrix Cauchy--Schwarz inequality,
\begin{align*}
\E[aa^\top]
&\succeq
\E[ab^\top]
\E[bb^\top]^{-1}
\E[ba^\top] \\
&=
(U^\top S U)^{-1}.
\end{align*}
Taking the trace,
\begin{align*}
\E\big[\|U^\top(\hat{x}-x)\|_2^2\big]
\ge
\Tr\big((U^\top S U)^{-1}\big).
\end{align*}
Since $U$ has orthonormal columns,
\begin{align*}
\|\hat{x}-x\|_2^2
\ge
\|U^\top(\hat{x}-x)\|_2^2,
\end{align*}
and therefore
\begin{align}
\E\big[\|\hat{x}-x\|_2^2/d\big]
\ge
\frac{1}{d}\Tr\big((U^\top S U)^{-1}\big).
\label{eq:direct_noise_free}
\end{align}

\end{proofEnd}

Corollary~\ref{thm:m_bound2} gives a non-trivial bound even when no noise is added to the encoder. Even when the encoder is deterministic, the rank of the \emph{expected} FIM over the prior limits the adversary's reconstruction accuracy. Specifically, the attack becomes harder as $\E\big[G\big]$ becomes lower-rank ($r < d$).
This matches our intuition: if the encoder captures information only along a consistent, low-dimensional subspace and discards the rest, the privacy of the input will enhance.

The bound also shows that reconstruction is easier (\emph{i.e.}, the bound is smaller) when the null space contains directions along which $S$ has large eigenvalues.
This again matches our intuition: even if the encoder discards directions where $S$ has large eigenvalues (\emph{i.e.}, directions that the prior already constrains tightly), such directions are easier to guess from the prior.
To summarize, our new bounds show that reconstruction is harder when: (1) the encoder discards more informative directions, and (2) the discarded directions are ones that are hard to guess from the prior (small eigenvalues in $S$).
%

%

\section{Extending to Metrics Beyond MSE}
\label{sec:psi_bound}

Next, we discuss how the bounds from Section~\ref{sec:m_bound} can be extended to other popular metrics that are considered to better capture the semantic similarity between data points.
Specifically, we consider norm-based metrics, where the input is projected into some high-dimensional space via a function $\Psi(x)$, and similarity is measured by the (squared) $\ell_2$-norm in that space, \emph{i.e.}, $\|\Psi(x_1) - \Psi(x_2)\|^2$.
This general form covers a wide range of popular metrics: when comparing two images, popular metrics such as LPIPS~\cite{lpips} and  CLIP~\cite{clip}; and in the data retrieval literature, vector databases~\cite{rag} often measure relatedness between data points using the $\ell_2$ distance between their representations in a high-dimensional embedding space. This section applies broadly to this large family of $\ell_2$-based similarity metrics.

\subsection{Bounding the $\ell_2$-Norm for Function $\Psi$}
\label{sec:psi_bound1}

First, we derive a bound for a general function $\Psi$, when the encoder has randomness.
Our bound is an extension of the generalized form of the multivariate van Trees inequality from Gill \& Levit~\cite{gill1995applications}, restated in Appendix~\ref{app:vantrees_gill1995}, and proof is in Appendix~\ref{sec:proofs}.


\begin{theoremEnd}[proof end]{theorem}[$\Psi$-bound]\label{thm:psi_bound1}
Let $\Psi : \R^{d} \to \R^{k}$ be a twice-differentiable function, with
$J_\Psi = \partial \Psi / \partial x$. Let $P_{1}$ and $P_{2}$ be two
orthogonal projectors that are orthogonal to each other and
$P_{1} + P_{2} = I_d$ (it can be generalized to arbitrary number of
projectors, but we keep it two for simplicity). Define for
$g \in \{1, 2\}$:
\[
n_g = \E\big[\big\|J_\Psi P_g\big\|_F^2\big],
\qquad
t_g = \Delta_g\Psi + J_\Psi P_g s_\pi,
\]
where $\Psi_i : \R^d \to \R$ is the $i$-th component of $\Psi$, and the $i$-th row of $\Delta_g\Psi$ is
$(\Delta_g \Psi)_i = \Tr\big(P_g\, \nabla^2 \Psi_i\big)$. Next, we define
a $2 \times 2$ matrix $D$, with the following $(i, j)$-th elements:
\[
D_{ij} = \frac{1}{\sigma^2}\E\Big[
  \langle J_{\mathrm{Enc}_{\mathrm{D}}}P_i J_\Psi^\top,\,
          J_{\mathrm{Enc}_{\mathrm{D}}}P_j J_\Psi^\top \rangle_F \Big]
  + \E\big[\big\langle t_{i},\, t_{j}\big\rangle\big],
\]
where $\langle \cdot, \cdot \rangle_F$ is a Frobenius inner product.
Suppose Assumption~\ref{asm:standing} holds, that
$J_\Psi(x) P_g\, p(e, x) \to 0$ as $\|x\| \to \infty$ for a.e.\ $e$ and
each $g$, that the moments $n_g$ and $D_{ij}$ are finite, and that
$D \succ 0$. Then for any attacker $\hat x = \mathrm{Att}(e)$,
\begin{equation}\label{eq:oplus}
\E\,\|\Psi(\hat x) - \Psi(x)\|^2 \;\ge\;
n^\top D^{-1} n,
\qquad n = [n_{1} \;\; n_{2}]^\top.
\end{equation}
\end{theoremEnd}

\begin{proofEnd}
Apply Lemma~\ref{lem:gill-levit} with
\[
C(x) \;=\; \sum_{g \in \{1, 2\}} w_g\, J_\Psi(x)\, P_g,
\qquad w_g \in \R.
\]
for arbitrary $w_1, w_2 \in \R$. Since $P_g$ is a projector, the numerator of Lemma~\ref{lem:gill-levit} becomes
\begin{align*}
\E\big[\Tr(J_\Psi C^\top)\big] &= 
\sum_{g \in \{1, 2\}} w_g \E\big[\Tr(J_\Psi(x) P_g J_\Psi(x)^\top)\big]\\
&= \sum_{g \in \{1, 2\}} w_g \E\big[\Tr(J_\Psi(x) P_g P_g J_\Psi(x)^\top)\big]\\ 
&=
\sum_{g \in \{1, 2\}} w_g \E\big[\Tr(P_g J_\Psi(x)^\top J_\Psi(x) P_g) \big]\\
&=
\E\big[\sum_{g \in \{1, 2\}} w_g \|J_\Psi(x) P_g\|_F^2 \big]\\ 
&= \sum_{g \in \{1, 2\}} w_g\, n_g = w^\top n, \qquad w = [w_1 \;\; w_2]^\top.
\end{align*}

Next, we consider the first denominator term of Lemma~\ref{lem:gill-levit}. The $i$-th row of $C$ is
$\sum_{g \in \{1, 2\}} w_g\nabla\Psi_i^\top P_g$, so
\begin{align*}
\E\big[&\Tr\big( C(x) \mathcal{I}_{e}(x)
    C(x)^{\top} \big) \big]\\
&=\frac{1}{\sigma^2}\E\big[\Tr\big( C(x) J_{\mathrm{Enc}_D}(x)^\top J_{\mathrm{Enc}_D}(x)
    C(x)^{\top} \big) \big]\\
&=\frac{1}{\sigma^2}\E\Big[\big\|J_\mathrm{Enc_{D}} C^\top\big\|_F^2\Big]\\
&= \frac{1}{\sigma^2}\E\Big[\sum_i\big\|J_\mathrm{Enc_{D}} \big(\sum_{g \in \{1, 2\}} w_g\nabla\Psi_i^\top P_g\big)^\top\big\|_F^2\Big]\\
&= \frac{1}{\sigma^2}\E\Big[\sum_i\big\| \sum_{g \in \{1, 2\}} w_g J_\mathrm{Enc_{D}}  P_g\nabla\Psi_i \big\|_F^2\Big].
\end{align*}
For notational simplicity, if we let $K_g = J_\mathrm{Enc_{D}} P_g J_\Psi^\top$:
\begin{align*}
&= \frac{1}{\sigma^2}\E\Big[\sum_i\big\langle \sum_{g \in \{1, 2\}} w_g J_\mathrm{Enc_{D}}  P_g\nabla\Psi_i, \sum_{g \in \{1, 2\}} w_g J_\mathrm{Enc_{D}}  P_g\nabla\Psi_i  \big\rangle\Big]\\
&= \frac{1}{\sigma^2}w^\top 
\begin{pmatrix}
\E\big[ \langle K_1, K_1\rangle_F \big] & \E\big[ \langle K_1, K_2\rangle_F \big] \\
\E\big[ \langle K_2, K_1\rangle_F \big] & \E\big[ \langle K_2, K_2\rangle_F \big]
\end{pmatrix}
w.
\end{align*}

Finally, let's consider the second denominator term of Lemma~\ref{lem:gill-levit}. The $i$-th row of the divergence of $C$ is:
\begin{align*}
\divg\big(\sum_{g \in \{1, 2\}} w_g\nabla\Psi_i^\top P_g\big) &= \sum_{g \in \{1, 2\}} w_g\divg\big(\nabla\Psi_i^\top P_g\big)\\
&= \sum_{g \in \{1, 2\}} w_g\nabla\!\cdot\! \big(P_g \nabla\Psi_i\big)\\ &= \sum_{g \in \{1, 2\}} w_g \Tr\big(P_g \nabla^2\Psi_i\big)\\
&= \sum_{g \in \{1, 2\}} w_g (\Delta_G\Psi)_i.
\end{align*}
Hence, 
\begin{align*}
\E\big[\|\divg C + C s_\pi\|^2\big] &= \E\big[\|\sum_{g \in \{1, 2\}} w_g\, t_g\|^2\big]\\
&= 
\sum_{g \in \{1, 2\}} \sum_{g' \in \{1, 2\}} w_g w_{g'} \langle t_g, t_{g'} \rangle\\ &= w^\top 
\begin{pmatrix}
\E[\big\langle t_{1},\, t_{1}\big\rangle] & \E[\big\langle t_{1},\, t_{2}\big\rangle] \\
\E[\big\langle t_{2},\, t_{1}\big\rangle] & \E[\big\langle t_{2},\, t_{2}\big\rangle]
\end{pmatrix}
w.
\end{align*}
Substituting all the components back into Lemma~\ref{lem:gill-levit}, every nonzero $w$ gives a valid bound
$(w^\top n)^2 / (w^\top D w)$. To get its supremum, we define an inner product $\langle u, v \rangle_{D} = u^\top D v$ ($D \succ 0$ required). Then, $w^\top n = \langle w, D^{-1}n \rangle_{D}$. From Cauchy-Schwarz:
\[
 (w^\top n)^2 \le \langle w, w \rangle_{D} \langle D^{-1}n, D^{-1}n \rangle_{D}
 = (w^\top D w)(n^\top D^{-1} n).
\]
The final step leverages the fact that $D$ is symmetric.
Thus,
\[
\sup_{w \ne 0} \frac{(w^\top n)^2}{w^\top D w} \;=\; n^\top D^{-1} n,
\]
with equality at $w \propto D^{-1}n$.
\end{proofEnd}
%
%
The LPIPS bound may be hard to interpret. To make the interpretation easier, consider the case where $P_1$ and $P_2$ are chosen to be the projectors onto the range space and null space of $\E\big[G\big]$, respectively (call them $P_{\mathrm{r}}$ and $P_{\mathrm{n}}$, with their respective $t$ and $n$ vectors denoted $t_{\mathrm{r}}$, $t_{\mathrm{n}}$, $n_{\mathrm{r}}$, and $n_{\mathrm{n}}$; note that $\E\big[G\big]$ is symmetric, and $\mathrm{null}(\E\big[G\big]^\top)=\mathrm{null}(\E\big[G\big])$).
Then $J_\mathrm{Enc_{D}} P_{\mathrm{r}}=J_\mathrm{Enc_{D}}$ and $J_\mathrm{Enc_{D}} P_{\mathrm{n}}=0$ for $\pi$-almost every $x$, and:
\[
D \;=\; 
\begin{pmatrix}
\frac{1}{\sigma^2}\E\Big[ \|J_\mathrm{Enc_{D}} J_\Psi^\top\|^2_F \Big] + \E\big[\big\langle t_{\mathrm{r}},\, t_{\mathrm{r}}\big\rangle\big] & \E\big[\big\langle t_{\mathrm{r}},\, t_{\mathrm{n}}\big\rangle\big] \\
\E\big[\big\langle t_{\mathrm{n}},\, t_{\mathrm{r}}\big\rangle\big] & \E\big[\big\langle t_{\mathrm{n}},\, t_{\mathrm{n}}\big\rangle\big]
\end{pmatrix}.
\]
When the encoder discards exactly the directions that $\Psi$ cares about (\emph{i.e.}, $J_\Psi$ and $P_{\mathrm{n}}$ are well aligned), $n_{\text{n}}$ is large while $n_{\text{r}}$ is small, and the dominating term in $n^\top D^{-1} n$ becomes $n_{\text{n}}^2 / \E\big[\big\langle t_{\mathrm{n}},\, t_{\mathrm{n}}\big\rangle\big]$, which is finite and does \emph{not} scale with $1/\sigma^2$. In other words, the bound stays large no matter how noisy the encoder is---because the encoder's spectral geometry discards the very information that $\Psi$ needs anyway.

If the encoder instead preserves the directions that $\Psi$ captures (\emph{i.e.}, $J_\Psi$ and $P_{\mathrm{r}}$ are well aligned), $n_{\text{r}}$ is large while $n_{\text{n}}$ is small. Now the term with $1/\sigma^2$ dominates.
This makes the bound driven primarily by the noise in the encoder, and the bound collapses as the noise approaches zero.
Again, this is intuitive---since the encoder preserves the information relevant to $\Psi$, a large amount of noise must be added to prevent precise reconstruction under the $\Psi$-metric.



\subsection{Extending with Zero Noise}
\label{sec:psi_bound2}

Next, we extend the $\Psi$-bound to the case where no noise is added to the encoder output ($\sigma=0$), analogous to Corollary~\ref{thm:m_bound2}.
The bound is nontrivial only when $\E\big[G\big]$ is not full rank.

\begin{theoremEnd}[proof end]{corollary}[Zero-noise $\Psi$-bound]\label{thm:psi_bound2}
Let $P_\mathrm{n}$ be the projector onto the null space
$\mathrm{null}(\E[G])$, and suppose $\E[G]$ is not full rank. Let $t_\mathrm{n}$ and $n_\mathrm{n}$ be the $t$ and $n$ vectors for $P_\mathrm{n}$, as defined in Theorem~\ref{thm:psi_bound1}. Then, for any attack
$\hat{x}=\mathrm{Att}(\mathrm{Enc_D}(x))$ on a deterministic encoder $\mathrm{Enc_D}(x)$,
\begin{align}
\E\big[\|\Psi(\hat{x})-\Psi(x)\|^2\big]
&\ge
\frac{n_\mathrm{n}^2}
{\E\big[\|t_\mathrm{n}\|^2\big]} \\
&=
\frac{
\big(\E\big[\|J_\Psi P_\mathrm{n}\|_F^2\big]\big)^2
}{
\E\big[
\big\|\Delta_\mathrm{n}\Psi
+
J_\Psi P_\mathrm{n}s_\pi\big\|^2
\big]
}.
\label{eq:psi_bound_zero_noise}
\end{align}
When $\E[G]$ is full rank, there is no null-space component, and the bound becomes zero.
\end{theoremEnd}

\begin{proofEnd}
Let $U\in\R^{d\times(d-r)}$ be an orthonormal basis for
$\mathrm{null}(\E[G])$, where $r=\mathrm{rank}(\E[G])$, so that
$P_\mathrm{n}=UU^\top$.
As shown in the proof of Corollary~\ref{thm:m_bound2},
$J_{\mathrm{Enc_D}}(x)U=0$ for every $x$, and therefore
$\mathrm{Enc_D}(x)$ is invariant along the directions spanned by $U$.
Again, let $R$ be an orthonormal basis for $\mathrm{range}(\E[G])$, and define
$v=R^\top x$ and $z=U^\top x$. Then, $\mathrm{Enc_D}(Rv+Uz)$ does not
depend on $z$, and any attack
$\hat{x}=\mathrm{Att}(\mathrm{Enc_D}(x))$, $\Psi(\hat{x})$ also does not
depend on $z$.

Since $P_\mathrm{n}=UU^\top$,
\begin{align*}
(\Delta_\mathrm{n}\Psi)_i
&=
\Tr\big(P_\mathrm{n}\nabla^2\Psi_i\big) \\
&=
\Tr\big(UU^\top\nabla^2\Psi_i\big) \\
&=
\Tr\big(U^\top\nabla^2\Psi_i U\big),
\end{align*}
where the last equality follows from the cyclic invariance of the trace.
Writing $U=[u_1,\ldots,u_{d-r}]$, we have
\begin{align*}
\Tr\big(U^\top\nabla^2\Psi_i U\big)
&=
\sum_{j=1}^{d-r}
u_j^\top\nabla^2\Psi_i\,u_j.
\end{align*}
Since $z_j=u_j^\top x$, differentiation with respect to $z_j$ corresponds
to differentiation along the direction $u_j$, and therefore
\begin{align*}
\frac{\partial^2\Psi_i}{\partial z_j^2}
=
u_j^\top\nabla^2\Psi_i\,u_j.
\end{align*}
Hence,
\begin{align*}
(\Delta_\mathrm{n}\Psi)_i
=
\sum_{j=1}^{d-r}
\frac{\partial^2\Psi_i}{\partial z_j^2}.
\end{align*}
Also,
\begin{align*}
J_\Psi P_\mathrm{n}s_\pi
&=
J_\Psi U U^\top s_\pi.
\end{align*}
Writing $U=[u_1,\ldots,u_{d-r}]$ and recalling that
$z_j=u_j^\top x$, the $(i,j)$-th entry of $J_\Psi U$ is
\begin{align*}
(J_\Psi U)_{ij}
&=
\nabla_x\Psi_i(x)^\top u_j \\
&=
\frac{\partial\Psi_i}{\partial z_j}.
\end{align*}
Thus, $J_\Psi U$ is the Jacobian of $\Psi$ with respect to the null-space
coordinates $z$.
Again from the proof of Corollary~\ref{thm:m_bound2}, 
\begin{align*}
\big(U^\top s_\pi(x)\big)_j
&=
\frac{\partial\log\pi(v,z)}{\partial z_j}.
\end{align*}
Therefore, taking the $i$-th component of
\begin{align*}
J_\Psi P_\mathrm{n}s_\pi
&=
(J_\Psi U)(U^\top s_\pi),
\end{align*}
gives
\begin{align*}
\big(J_\Psi P_\mathrm{n}s_\pi\big)_i
&=
\sum_{j=1}^{d-r}
(J_\Psi U)_{ij}
\big(U^\top s_\pi\big)_j \\
&=
\sum_{j=1}^{d-r}
\frac{\partial\Psi_i}{\partial z_j}
\frac{\partial\log\pi(v,z)}{\partial z_j}.
\end{align*}

Now, recall that
\begin{align*}
t_\mathrm{n}
=
\Delta_\mathrm{n}\Psi
+
J_\Psi P_\mathrm{n}s_\pi.
\end{align*}
For each component $i$,
\begin{align*}
(t_\mathrm{n})_i
&=
(\Delta_\mathrm{n}\Psi)_i
+
\big(J_\Psi P_\mathrm{n}s_\pi\big)_i.
\end{align*}
Using the expressions derived above gives, 
\begin{align*}
(t_\mathrm{n})_i
&=
\sum_{j=1}^{d-r}
\frac{\partial^2\Psi_i}{\partial z_j^2}
+
\sum_{j=1}^{d-r}
\frac{\partial\Psi_i}{\partial z_j}
\frac{\partial\log\pi(v,z)}{\partial z_j} \\
&=
\sum_{j=1}^{d-r}
\left(
\frac{\partial^2\Psi_i}{\partial z_j^2}
+
\frac{\partial\Psi_i}{\partial z_j}
\frac{\partial\log\pi(v,z)}{\partial z_j}
\right).
\end{align*}
Since
\begin{align*}
\frac{\partial\log\pi(v,z)}{\partial z_j}
=
\frac{1}{\pi(v,z)}
\frac{\partial\pi(v,z)}{\partial z_j},
\end{align*}
we have
\begin{align*}
(t_\mathrm{n})_i
&=
\sum_{j=1}^{d-r}
\left(
\frac{\partial^2\Psi_i}{\partial z_j^2}
+
\frac{\partial\Psi_i}{\partial z_j}
\frac{1}{\pi(v,z)}
\frac{\partial\pi(v,z)}{\partial z_j}
\right) \\
&=
\frac{1}{\pi(v,z)}
\sum_{j=1}^{d-r}
\left(
\pi(v,z)
\frac{\partial^2\Psi_i}{\partial z_j^2}
+
\frac{\partial\pi(v,z)}{\partial z_j}
\frac{\partial\Psi_i}{\partial z_j}
\right).
\end{align*}
By the product rule,
\begin{align*}
\frac{\partial}{\partial z_j}
\left(
\pi(v,z)
\frac{\partial\Psi_i}{\partial z_j}
\right)
&=
\frac{\partial\pi(v,z)}{\partial z_j}
\frac{\partial\Psi_i}{\partial z_j}
+
\pi(v,z)
\frac{\partial^2\Psi_i}{\partial z_j^2}.
\end{align*}
Therefore,
\begin{align*}
(t_\mathrm{n})_i
&=
\frac{1}{\pi(v,z)}
\sum_{j=1}^{d-r}
\frac{\partial}{\partial z_j}
\left(
\pi(v,z)
\frac{\partial\Psi_i}{\partial z_j}
\right).
\end{align*}

Since $\mathrm{Enc_D}(Rv+Uz)$ is independent of $z$, the attack output is
also independent of $z$. We therefore write
\begin{align*}
\hat{x}(v)
=
\mathrm{Att}\big(\mathrm{Enc_D}(Rv+Uz)\big),
\end{align*}
where the right-hand side is the same for every $z$. Define
$h:\R^r\times\R^{d-r}\rightarrow\R^{\dim(\Psi)}$ by
\begin{align*}
h(v,z)
&=
\Psi(\hat{x}(v))
-
\Psi(Rv+Uz),
\end{align*}
and let
\begin{align*}
h_i(v,z)
&=
\Psi_i(\hat{x}(v))
-
\Psi_i(Rv+Uz)
\end{align*}
denote its $i$-th component.
Then,
\begin{align*}
&\E_{(v,z)\sim\pi}
\big[
h_i \cdot (t_\mathrm{n})_i
\big]\\
&=
\int
h_i(v,z)(t_\mathrm{n})_i
\pi(v,z)
\,dv\,dz \\
&=
\int
h_i(v,z)
\left[
\frac{1}{\pi(v,z)}
\sum_{j=1}^{d-r}
\frac{\partial}{\partial z_j}
\left(
\pi(v,z)
\frac{\partial\Psi_i}{\partial z_j}
\right)
\right]
\pi(v,z)
\,dv\,dz \\
&=
\sum_{j=1}^{d-r}
\int
h_i(v,z)
\frac{\partial}{\partial z_j}
\left(
\pi(v,z)
\frac{\partial\Psi_i}{\partial z_j}
\right)
\,dv\,dz.
\end{align*}
Applying integration by parts with respect to $z_j$ gives
\begin{align*}
&\int
h_i(v,z)
\frac{\partial}{\partial z_j}
\left(
\pi(v,z)
\frac{\partial\Psi_i}{\partial z_j}
\right)
\,dv\,dz \\
&=
\int
\left[
h_i(v,z)
\pi(v,z)
\frac{\partial\Psi_i}{\partial z_j}
\right]_{z_j=-\infty}^{z_j=\infty}
\,dv\,dz_{-j} \\
&\quad
-
\int
\frac{\partial h_i(v,z)}{\partial z_j}
\pi(v,z)
\frac{\partial\Psi_i}{\partial z_j}
\,dv\,dz,
\end{align*}
where $dz_{-j}$ denotes integration over all coordinates of $z$ other
than $z_j$. The first term vanishes by the tail condition. Hence,
\begin{align*}
\E_{(v,z)\sim\pi}
\big[
h_i \cdot (t_\mathrm{n})_i
\big]
&=
-
\sum_{j=1}^{d-r}
\int
\frac{\partial h_i(v,z)}{\partial z_j}
\pi(v,z)
\frac{\partial\Psi_i}{\partial z_j}
\,dv\,dz.
\end{align*}
Finally, since $\hat{x}(v)$ is independent of $z$,
\begin{align*}
\frac{\partial h_i(v,z)}{\partial z_j}
&=
-
\frac{\partial}{\partial z_j}
\Psi_i(Rv+Uz) =
-
\frac{\partial\Psi_i}{\partial z_j},
\end{align*}
and
\begin{align*}
\E_{(v,z)\sim\pi}
\big[
h_i \cdot (t_\mathrm{n})_i
\big]
&=
\E_{(v,z)\sim\pi}\left[
\sum_{j=1}^{d-r}
\left(
\frac{\partial\Psi_i}{\partial z_j}
\right)^2
\right].
\end{align*}
Summing over $i$ and dropping the $_{(v,z)\sim\pi}$ subscript for simplicity gives
\begin{align*}
\E\big[\langle h,t_\mathrm{n}\rangle\big]
&=
\E\big[\|J_\Psi U\|_F^2\big] = \E\big[\|J_\Psi UU^\top\|_F^2\big] \\
&=
\E\big[\|J_\Psi P_\mathrm{n}\|_F^2\big] =
n_\mathrm{n},
\end{align*}
where the second equality is due to $U$ having orthonormal columns.
Finally, by Cauchy--Schwarz,
\begin{align*}
n_\mathrm{n}^2
&=
\Big(
\E\big[\langle h,t_\mathrm{n}\rangle\big]
\Big)^2 \\
&\le
\E\big[\|h\|^2\big]
\E\big[\|t_\mathrm{n}\|^2\big].
\end{align*}
Substituting $h=\Psi(\hat{x})-\Psi(x)$ and rearranging gives
\begin{align*}
\E\big[\|\Psi(\hat{x})-\Psi(x)\|^2\big]
\ge
\frac{n_\mathrm{n}^2}
{\E\big[\|t_\mathrm{n}\|^2\big]},
\end{align*}
which proves the result.
\end{proofEnd}
Again, the numerator $n_{\text{n}} = \E\big[\|J_\Psi P_{\text{n}}\|_F^2\big]$ measures how sensitive $\Psi$ is to the directions the encoder discards. A large numerator means the encoder is discarding the directions that $\Psi$ cares, and hence privacy improves (bound goes up). If the encoder instead preserves the directions that $\Psi$ cares, the numerator shrinks and the bound goes down. When $\E\big[G\big]$ is full rank ($r=d$), the bound becomes zero.
The lesson from this bound is similar to that of Theorem~\ref{thm:psi_bound1}: a privacy-enhancing encoder with respect to $\Psi$ should
(1) have low rank ($r \ll d$), \emph{i.e.}, discard as many informative directions as possible, and, especially, (2) discard the directions that $\Psi$ cares about.
\section{Practical Considerations}
\label{sec:practical_considerations}

In this section, we discuss several practical considerations for calculating and applying the bound in real-world use cases.

\subsection{Choosing $P_1$ and $P_2$}
\label{sec:heuristics_projector_choice}

In Theorem~\ref{thm:psi_bound1}, any choice of orthogonal projectors $P_1$ and $P_2$ (or even more than two projectors) gives a valid bound, though different choices yield different tightness. Throughout the rest of the paper, we choose $P_1 = P_{\mathrm{r}}$ and $P_2 = P_{\mathrm{n}}$---the projectors onto the range space and null space of $\E\big[G\big]$, respectively. This choice is easy to interpret and connects well to Corollary~\ref{thm:psi_bound2}, but it may not yield an optimal bound.
As our evaluation will show, our LPIPS bound is not always tight, and a tighter bound may be achievable by choosing a different set of projectors (likely more than two).
We leave such improvements to future work.

\subsection{Estimating $S$}
\label{sec:heuristics_s_estimation}

To use the new bounds, one must calculate $S=\E_\pi\!\big[s_\pi(x) s_\pi(x)^\top\big]$, where $s_\pi(x)=\nabla_x\log\pi(x)$ is the score of each input $x$.
For image datasets, Maeng et al.~\cite{maeng2023bounding} used a technique called sliced score matching~\cite{sliced_score_matching} to train a model called NICE~\cite{nice}, which predicts $s_\pi(x)$ from $x$.
Instead, we show that a well-trained, off-the-shelf diffusion model (in particular, DDPM~\cite{ddpm}) can be used to predict $s_\pi(x)$.
We show in Section~\ref{sec:eval_S_estimation} that our method provides a better estimate than a trained NICE~\cite{nice} model; this is an additional contribution of our work.

%
%
%

DDPM starts from a maximally noisy image $x_T$ and denoises it back to $x_{T-1}$, $x_{T-2}, \ldots, x_0$, where $x_t$ is the noisy image at step $t$ and $x_0$ is the original, noise-free image.
A key component of DDPM is training a model $\epsilon_\theta(x_t, t)$, which predicts the noise present in the image at step $t$.
Song et al.~\cite{song2020score} showed that there is a relationship between $\epsilon_\theta(x_t, t)$ and $s_{\pi}(x)$; specifically, $s_{\pi_t}(x_t) = -\epsilon_\theta(x_t, t)/\sqrt{1-\bar{\alpha}_t}$, where $\bar{\alpha}_t$ is a coefficient determined by the noise schedule (provided with the pretrained model), and $\pi_t$ is the distribution of $x_t$.

Recall that we operate on a smoothed (\emph{i.e.}, slightly noisy) image $x = x_0 + \mathcal{N}(0, \sigma_s^2 I_d)$, for some small $\sigma_s$.
In DDPM, $x_t = \sqrt{\bar{\alpha}_t}\big(x_0 + \sqrt{(1 - \bar{\alpha}_t)/\bar{\alpha}_t}\,z\big)$, where $z \sim \mathcal{N}(0, I_d)$.
If we choose $t$ such that $\sqrt{(1 - \bar{\alpha}_t)/\bar{\alpha}_t} = \sigma_s$, then $x_t = \sqrt{\bar{\alpha}_t}\,x$, and
\[
s_\pi(x) = \sqrt{\bar{\alpha}_t}\,s_{\pi_t}(x_t) = -\sqrt{\bar{\alpha}_t/(1-\bar{\alpha}_t)}\,\epsilon_\theta\big(\sqrt{\bar{\alpha}_t}\,x,\, t\big).
\]
This allows us to calculate the score function $s_\pi(x)$ using a trained DDPM.

Throughout our evaluation, we download publicly available, well-trained DDPMs (for MNIST~\cite{ddpm_mnist} and CIFAR-10~\cite{google_ddpm_cifar10}) and use them to estimate $s_\pi(x)$.
Then, $S=\E\!\big[s_\pi(x) s_\pi(x)^\top\big]$ is calculated by averaging over all images in the training set.
Note that DDPM assumes pixel values in $[-1, 1]$, so a scaling factor of 2 must be applied when using pixel values in $[0, 1]$.

\subsection{Estimating $\E\big[G\big]$}
\label{sec:heuristics_g_estimation}

Calculating the bounds requires (1) estimating $\E\big[G\big]$, and (2) separating its range space and null space.
Correctly estimating $\E\big[G\big]$ is required for all four bounds. Correctly separating its null space and range space is needed for the noiseless bounds (Corollary~\ref{thm:m_bound2} and Corollary~\ref{thm:psi_bound2}).

\textbf{Estimating $\E\big[G\big]$.}
Estimating $\E\big[G\big]$ over $\pi$ requires calculating $J_{\mathrm{Enc_{D}}}(x)$ and averaging over $x \sim \pi$.
For efficiency, we sample $k$ images from the training data and average over the sampled images.
Ideally, using the entire training set gives the most accurate estimate of the bound, but we empirically find that a smaller $k$ suffices in most cases.
The effect of the choice of $k$ is further evaluated in Section~\ref{sec:eval_sensitivity_eg_k}.

\textbf{Splitting range and null space.}
The bounds from Corollary~\ref{thm:m_bound2} and Corollary~\ref{thm:psi_bound2} (the bounds for $\sigma=0$) require splitting the range space and null space of $\E\big[G\big]$.
While one can obtain the exact null space from $\E\big[G\big]$ once it is estimated, using this exact null space often gives a correct but conservative bound, because any subspace with a negligibly small but still nonzero eigenvalue is treated as part of the range space.
Alternatively, we explore setting a threshold $\tau$ and treating subspaces whose eigenvalues are smaller than $\tau \cdot \lambda_{\mathrm{max}}$ as null, where $\lambda_{\mathrm{max}}$ is the largest eigenvalue of $\E\big[G\big]$.
We empirically find that this gives a tighter bound. The effect of different choices of $\tau$ is evaluated in Section~\ref{sec:eval_sensitivity_eg_tau}.

\textbf{Calculating $J_{\mathrm{Enc_{D}}}(x)$.}
$J_{\mathrm{Enc_{D}}}(x)$ is an $m \times d$ matrix. Since privacy-enhancing encoders usually have a small output dimension $m$, the entire $J_{\mathrm{Enc_{D}}}(x)$ can typically be materialized at once in modern GPU memory.
When this is not possible, we instead calculate it column-by-column using a Jacobian-vector product (jvp).

\subsection{Estimating Other Terms in Theorem~\ref{thm:psi_bound1}}

Estimating $n_g = \E\big[\|J_\Psi P_g\|_F^2\big]$, $\E\big[\langle t_i, t_j\rangle\big]$, $\E\Big[ \|J_\mathrm{Enc_{D}} J_\Psi^\top\|^2_F \Big]$, and $\Tr\big(P_g \nabla^2 \Psi_i\big)$---the remaining terms of Theorem~\ref{thm:psi_bound1}---is even more expensive, because $\Psi$'s output dimension is large for LPIPS. Thus, we estimate these quantities through Rademacher probes~\cite{hutchinson1989stochastic}.
Throughout we use 16 probes per input, averaged over 8 inputs. We confirmed that increasing the number of probes beyond this provides diminishing return. Admittedly, averaging over more inputs may improve the bound further, but our LPIPS bounds are always valid with this approximated estimation.


\subsection{Bound as a Proxy for Invertibility}
\label{sec:ratio_heuristic}

As we will show in our evaluation, when the MSE or LPIPS bound is small, it is a strong indicator that the encoder's privacy-enhancing property is weak, \emph{i.e.}, that the original input can be reconstructed easily.
However, the absolute value of the bound becomes less informative as it grows and starts to saturate.
For example, for the MNIST dataset, an attacker can still achieve a reasonably low MSE or LPIPS by drawing white, squiggly, ``number-like'' lines on a black background, even when the attack fails to reconstruct any meaningful information.
This is a fundamental issue with norm-based similarity metrics---the meaning of their absolute value heavily depends on context. This makes using the absolute number of the bound ill-suited as a measure for privacy.

However, we empirically observe that the \emph{ratio} between the bound and its \emph{ceiling}---the bound that would be achieved if the encoder conveyed no information at all---is still a reasonable indicator of the encoder's privacy-enhancing properties.
In particular, we find that the ratio computed from the MSE bound works better, because this bound is tighter and less flat in the hard-to-reconstruct regions (as we later show in Section~\ref{sec:eval}).
From the observation, we define the following metric and call it the \emph{ratio to ceiling}, or \emph{RTC}:
\[
    \text{RTC} := \log \frac{\text{(ceiling of the MSE bound)}}{\text{(actual MSE bound)}},
\]
where $\log$ is the natural logarithm (base $e$).
The idea behind RTC follows popular privacy definitions that measure privacy as the (log of the) ratio between the adversary's belief before and after a privacy-leaking observation~\cite{pufferfish, inferential_privacy, bayesian_dp}.
Essentially, RTC measures how much the bound has shifted---\emph{i.e.}, how much easier reconstruction became---after the adversary observes the encoder's output, compared to guessing based on the prior $\pi$ alone.
A lower RTC indicates better privacy, and $\text{RTC}=0$ implies that the bound collapses to the prior-only bound (the ceiling).
We show in Section~\ref{sec:eval_rtc} that RTC serves as a reasonable empirical metric for privacy.

To summarize, our bounds serve two practical purposes: (1) when the absolute value of the bounded metric has direct interpretation (\emph{e.g.}, bounding the MSE of weight or height of a patient), the bound can directly limit them from being reconstructed; and (2) even when the bounded metric is only loosely indicative of the reconstruction quality (\emph{e.g.}, as in MSE for natural images), the \emph{ratio} between the bound and the ceiling of the bound (\emph{i.e.}, RTC) serves as a \emph{theoretically-principled} indicator for reconstruction hardness.
\begin{figure*}
    \centering
    \includegraphics[width=0.9\textwidth]{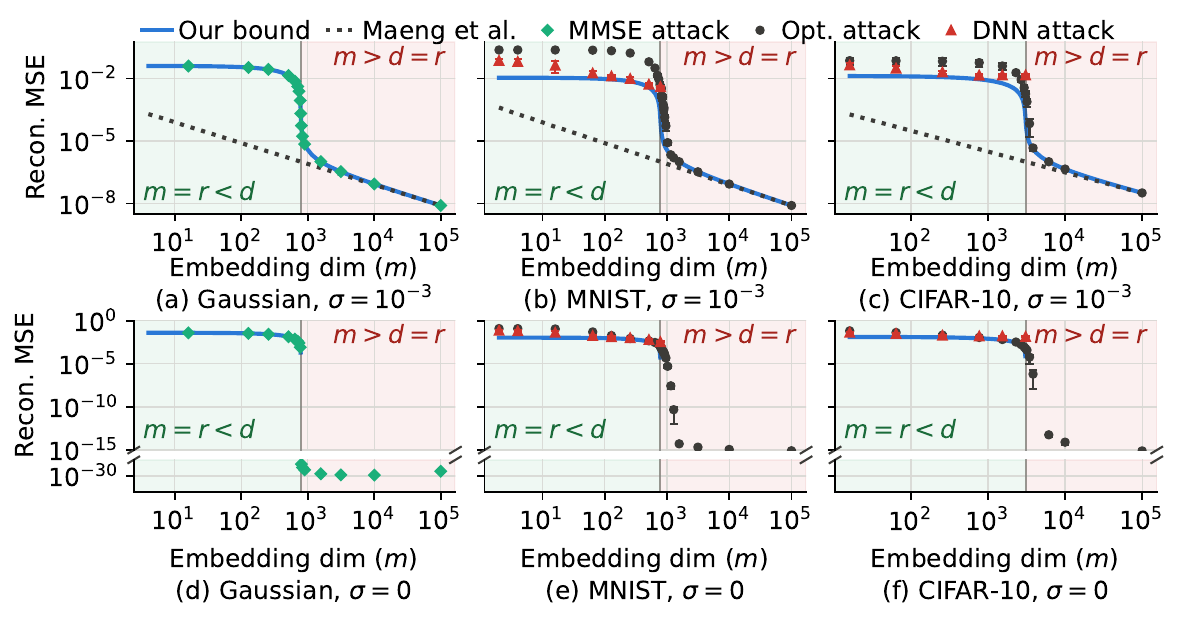}
    \caption{Our MSE bounds (solid blue line) closely lower-bound the attackers' achievable MSE for all the attackers. Bound from Maeng et al.~\cite{maeng2023bounding} (dotted line) becomes very loose when $r < d$ and is always zero when $\sigma=0$ (no encoder randomness). 
    }
    \label{fig:bound_vs_m}
\end{figure*}

\section{Evaluation}
\label{sec:eval}

Our evaluation aims to answer the following questions regarding our new bounds (Corollaries~\ref{thm:m_bound1}, \ref{thm:m_bound2}, and \ref{thm:psi_bound2}, and Theorem~\ref{thm:psi_bound1}):
\begin{itemize}[noitemsep, leftmargin=*, topsep=0pt]
    \item Do they work across attacks, encoders, and datasets?
    \item Are they tighter than the bound of Maeng et al.~\cite{maeng2023bounding}?
    \item How do the bounds change with the heuristics from Section~\ref{sec:practical_considerations}?
    \item Is RTC (Section~\ref{sec:ratio_heuristic}) indicative of reconstruction hardness?
\end{itemize}

\subsection{Evaluation Setup and Methodologies}
\label{sec:eval_setup}

\textbf{Datasets.} We evaluate our bounds on three datasets: (1) isotropic Gaussian noise (matching the dimensionality of MNIST for MSE and CIFAR-10 for LPIPS bound), sampled from $\mathcal{N}(0.5, 0.2^2I_d)$; (2) MNIST~\cite{mnist}; and (3) CIFAR-10~\cite{cifar10}. For MNIST and CIFAR-10, we add smoothing noise of $\mathcal{N}(0, 0.1^2I_{d})$, following Maeng et al.~\cite{maeng2023bounding}. For the Gaussian dataset, $S$ is calculated analytically (inverse of the covariance matrix). Others use DDPM to calculate $S$ (Section~\ref{sec:eval_S_estimation}).

\textbf{Encoders.} We test encoders with various architectures and weights.
The first are \emph{linear encoders} ($\mathrm{Enc}(x)=Wx+\mathcal{N}(0, \sigma^2I_d)$, $W \in \R^{m\times d}$) with varying output dimension $m$ and $\sigma \in \{0, 10^{-3}\}$ (small noise or no noise). The second are \emph{DNN encoders}, consisting of the first few layers of popular CNNs, followed by an additional compression block to control the output dimension $m$. The compression block consists of a compressing convolutional layer, followed by GELU and an optional AvgPool layer.
For MNIST, we use a simple 3-block CNN with three split points (early: after block~1; middle: after block 2; late: after block 3). For CIFAR-10, we use ResNet-18~\cite{resnet18_cifar10} with three split points (early: after the first convolutional layer; middle: after ResNet block 4; late: after ResNet block 6). To make the encoder differentiable, we replace MaxPool with AvgPool and ReLU with GELU.
The encoder is followed by a downstream model that consumes the embedding and performs the downstream task; this model consists of a decompression block followed by the remainder of the split DNN.
Details are in Appendix~\ref{app:arch}.

We study three different weights: (1) random weights, frozen after initialization;
(2) weights trained jointly with the downstream model on the target task;
and, for linear encoders only, (3) a weight matrix $W$ whose eigenvalues are analytically controlled to be $\lambda_i = d \cdot i^{-\alpha} / \sum_{j} j^{-\alpha}$, with a varying $\alpha$ to control the skewness.
This is achieved via $W = U\, \text{diag}(\sqrt{\lambda_i})\, V^\top$, with random unitary matrices $U$ and $V$.

\textbf{Attacks.} 
For reconstructing the Gaussian inputs $x \sim \mathcal{N}(\mu, \eta^2I_d)$ from the linear encoder $e = Wx + \mathcal{N}(0, \sigma^2I_m)$, we can analytically arrive to a minimum mean-squared error (MMSE) estimator, which is the posterior mean. For $\sigma > 0$,
\begin{equation}
    \hat{x} = \E\big[x | e\big] = \mu\1 + \big(\frac{W^\top W}{\sigma^2} + \frac{I_d}{\eta^2}\big)^{-1}\frac{W^\top(e - W\mu\1)}{\sigma^2}.
\label{eq:gaussian_mmse}
\end{equation}
For $\sigma=0$, 
\begin{equation}
    \hat{x} = \E\big[x | e\big] = \mu\1 + W^{+}(e - W\mu\1),
\label{eq:gaussian_mmse_zero_noise}
\end{equation}
where $W^+$ is a Moore–Penrose pseudoinverse of $W$ (proofs in Appendix~\ref{sec:proofs}).
We call this \emph{MMSE attack}. MMSE attacks are only defined for Gaussian inputs and linear encoders.

\begin{figure*}
    \centering
    \includegraphics[width=0.9\textwidth]{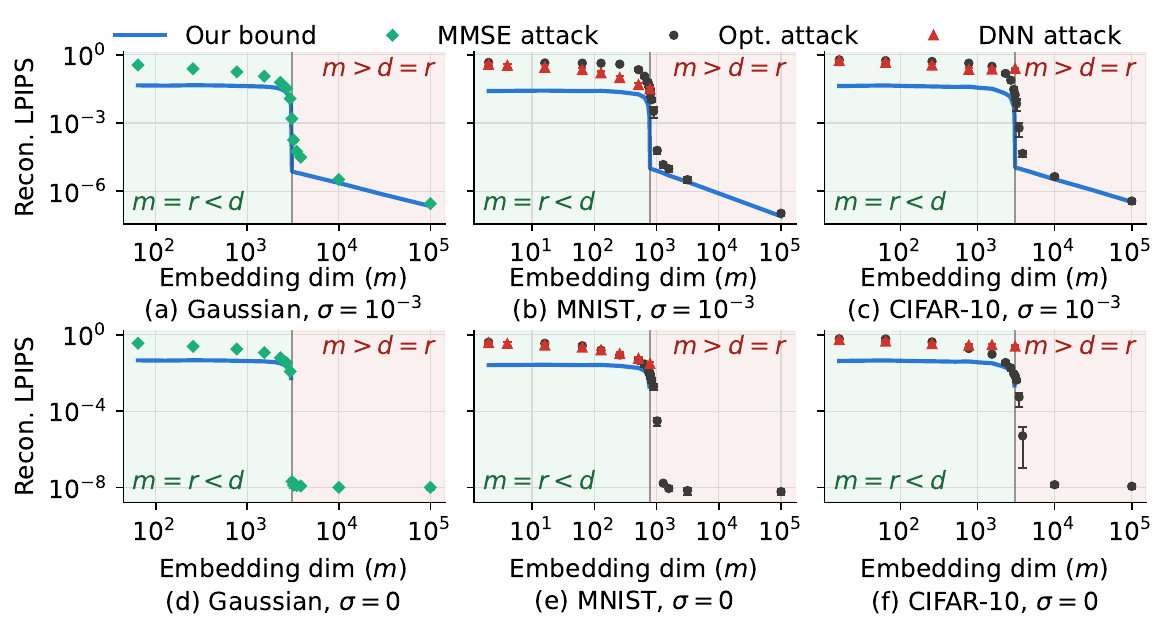}
    \caption{Our LPIPS bounds (solid blue line) are always met, and tight near $m \sim d$. For $m \ll d$, the bound is unfortunately looser and near-flat, being less informative than the MSE bounds. Still, this is the first working LPIPS bound to our knowledge.}
    \label{fig:bound_psi_vs_m}
\end{figure*}

For natural image datasets, we test two attacks. The first is an optimization-based attack (\emph{Opt. attack}), which minimizes a MAP-style objective with a tunable prior weight $\beta$:
\[
\hat{{x}}({e}) = \underset{{x^*}}{\arg\min}\;\|{e} - \mathrm{Enc}_D({x^*})\|_2^2 - \beta \log \pi({x}^*).
\]
For $\sigma > 0$, setting $\beta = 2\sigma^2$ recovers the exact maximum a posteriori (MAP) estimate. For $\sigma = 0$, $\beta$ is instead treated as a tunable hyperparameter.
For the data prior $\pi(x)$, we use the learned score from the pre-trained DDPM (Section~\ref{sec:heuristics_s_estimation}).
This generalizes existing optimization-based attacks that use heuristically designed proxies for the data prior~\cite{tv, dip}.
The second is a DNN-based attack (\emph{DNN attack}), which uses a trained DNN as the attacker, following the recipe of~\cite{ressfl}.
Details of the attacks are in Appendix~\ref{app:attack_params}.

\textbf{Additional design parameters.}
We calculate $S$ and $\E\big[G\big]$ exactly whenever possible. When we need to approximate, we use the following parameters unless stated otherwise:
when calculating $\E\big[G\big]$ with subsampled images, we use $k=256$ images; when approximating small-signal subspaces as the null space, we treat directions whose eigenvalues are smaller than 0.0001 times the largest eigenvalue ($\tau=10^{-4}$) as null. The effect of these choices is evaluated in Section~\ref{sec:eval_sensitivity}.
Finally, for LPIPS, we use the pretrained VGG-16 backbone with ReLU/MaxPool replaced by GELU/AvgPool (for twice-differentiability).
We confirme that this does not significantly affect the calculated LPIPS score.
For MNIST, we replicate the channel dimension, since LPIPS assumes three channels.

\subsection{Bounds with Linear Encoders}

We first evaluate our new bounds on linear encoders.
For linear encoders, the FIM is not data-dependent, so $\E\big[G\big]$ can be calculated exactly, and $r=\mathrm{rank}(\E\big[G\big])$ is simply $r=\min(d, m)$, provided the weights do not introduce additional null directions.
These two properties eliminate the need for many of the heuristics in Section~\ref{sec:heuristics_g_estimation}, isolating the effectiveness of the bound from potential errors introduced by the heuristics.


\subsubsection{Controlling the Rank}
First, we evaluate the bound with a random $W$, which makes $W$ isotropic, while controlling the rank.
%

\textbf{MSE bound.}
Figure~\ref{fig:bound_vs_m} plots our MSE bounds (blue solid line) from Corollary~\ref{thm:m_bound1} (when noise with $\sigma=10^{-3}$ is added; Figure~\ref{fig:bound_vs_m}(a--c)) and Corollary~\ref{thm:m_bound2} (when no noise is added; Figure~\ref{fig:bound_vs_m}(d--f)), along with the MSE actually achieved by the attacks.
For setups with noise $\sigma=10^{-3}$ (top row; a--c), we also plot the bound from Maeng et al.~\cite{maeng2023bounding} (Theorem~\ref{thm:2023_bound}; dotted line).
For setups with no noise (bottom row; d--f), the bound from Maeng et al.~\cite{maeng2023bounding} becomes zero.

The Gaussian MMSE attack exactly lands on top of our bounds. For natural images, Opt. attack performs better when $m > d$, while DNN attack performs better when $m < d$ (we only show the Opt. attack in the $m > d$ region). Still, our bound is valid for both attacks throughout, and is reasonably tight in most cases.
On the other hand, the bound from Maeng et al.~\cite{maeng2023bounding} (dotted line) is tight only when $d \ll m$ and $\sigma > 0$.
%
Again, this is because our bounds better account for the information lost through the encoder's spectral geometry.

We can also see that our bounds for noisy embeddings (Figure~\ref{fig:bound_vs_m}(a--c)) increase sharply around $m=r=d$. This is expected, since information starts to be lost through the encoder's spectral geometry once we cross the $m=r=d$ line.
Similarly, Figure~\ref{fig:bound_vs_m}(d--f) shows that the bound becomes nontrivial as this line is crossed, which is exactly when the encoder starts to lose information even without any added noise.

\begin{figure}[t]
    \centering
    \includegraphics[width=0.45\textwidth]{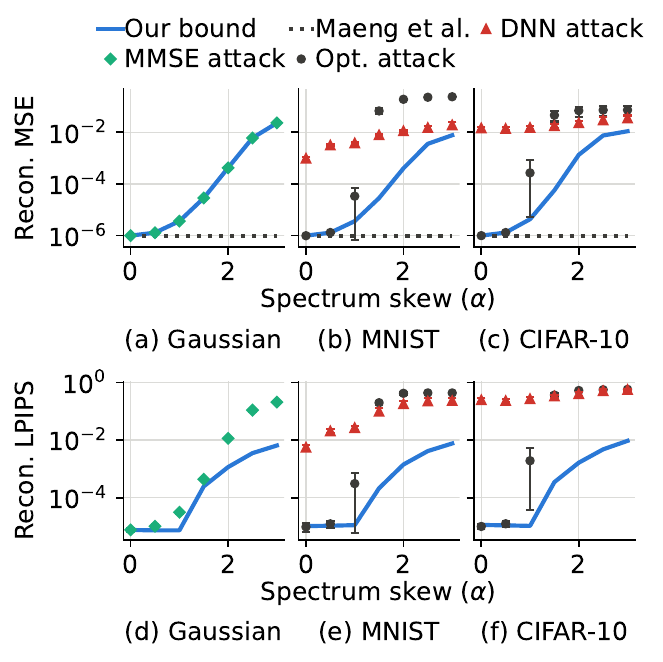}
    \caption{MSE and LPIPS bound with different skewness $\alpha$.}
    \label{fig:bound_vs_a}
\end{figure}

\textbf{LPIPS bound.}
Figure~\ref{fig:bound_psi_vs_m} shows the same plot using LPIPS as the similarity metric instead of MSE, based on Theorem~\ref{thm:psi_bound1} and Corollary~\ref{thm:psi_bound2}. The bounds are again always valid, but looser and nearly flat in the $m < d$ region, making them less informative than the MSE bound.
One potential reason for this, as explained in Section~\ref{sec:heuristics_projector_choice}, is that we simply choose the two projectors in Theorem~\ref{thm:psi_bound1} to be the projectors onto the range and null spaces, whereas a different (and larger) set of projectors could provide a tighter bound.
Again, we leave further improvement of the bound to future work.

\begin{figure}[t]
    \centering
    \includegraphics[width=0.45\textwidth]{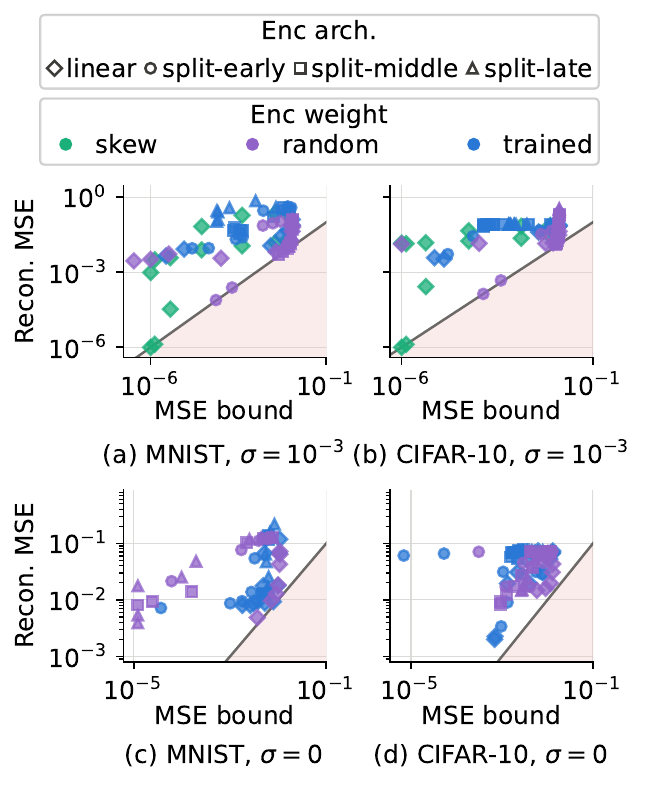}
    \caption{MSE bound vs. achieved MSE by the attacker.}
    \label{fig:bound_vs_attack_mse}
\end{figure}

\subsubsection{Controlling the Skewness of $\E\big[G\big]$}
Next, we keep $r=\mathrm{rank}(\E\big[G\big])$ constant at $m=r=d$ and vary the distribution of the eigenvalues using a controlled parameter $\alpha$, as discussed in Section~\ref{sec:eval_setup}.
We only show the case with small noise ($\sigma=10^{-3}$), since $\E\big[G\big]$ is full rank, and Corollary~\ref{thm:m_bound2} would give a trivial bound.
Again, the trend is similar: Opt. attack performs better for easier cases (lower $\alpha$), while DNN attack performs better for harder cases (higher $\alpha$). Still, our bounds hold for both attacks.

Figure~\ref{fig:bound_vs_a}(a--c) shows that our bounds hold for all the attacks. The bound remains tight for Gaussian data, but becomes looser on natural images with large $\alpha$ (but still tighter than the bound from Maeng et al.~\cite{maeng2023bounding}, which is the flat dotted line).
When the encoder is isotropic ($\alpha=0$), our bound coincides with that of Maeng et al.~\cite{maeng2023bounding}. However, as the spectrum becomes more skewed ($\alpha > 0$), our bound starts to increase, correctly capturing the fact that information loss along certain directions makes reconstruction harder.
The bound from Maeng et al.~\cite{maeng2023bounding}, on the other hand, stays flat.
%
Figure~\ref{fig:bound_vs_a}(d--f) again shows a similar trend: our bounds always hold against both attackers, but they are much looser than the MSE bound. 
%


\subsection{Bounds with DNN-based Encoders}

Next, we show results for general DNN-based encoders from Section~\ref{sec:eval_setup}. With DNN-based encoders, we do not have a straightforward way to control the rank or the skewness of $\E\big[G\big]$. We therefore simply plot the achieved attack MSE and LPIPS against the bounds for various encoder architectures (linear and DNN with early/middle/late splits) and weights (random, skew-controlled via $\alpha$ (only possible for linear encoders), and fully trained).
Here, we only plot the MSE and LPIPS from the Opt. attack, except for the linear encoders, whose DNN attackers were already trained from the previous section. This is because training DNN attackers for each encoder and takes too much time.

\begin{figure}[t]
    \centering
    \includegraphics[width=0.45\textwidth]{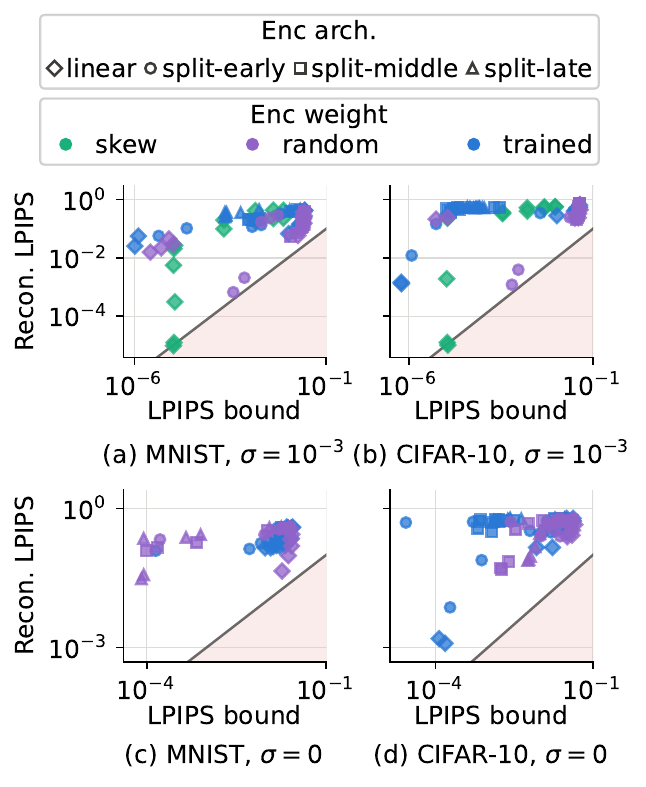}
    \caption{LPIPS bound vs. achieved LPIPS by the attacker.}
    \label{fig:bound_vs_attack_lpips}
\end{figure}

\textbf{MSE bound.}
Figures~\ref{fig:bound_vs_attack_mse}(a--b) plot the bound from Corollary~\ref{thm:m_bound1} against the attack MSE.
Across various encoder architectures and their weights, the bound always holds---no marker falls into the red-shaded region.
This shows that our bound (Corollary~\ref{thm:m_bound1}) holds reliably across various encoder architectures and training methods for their weights.

In general, bounds for linear or split-early encoders, especially with random or skewed weights, appear tighter. Meanwhile, bounds for split-middle and split-late encoders, especially with trained weights, appear looser, though still valid.
This matches empirical observations from prior work showing that reconstruction becomes harder as data passes through a deeper network~\cite{maeng2023bounding, privynet}. It may be that our bound still does not precisely capture this additional hardness due to encoder depth. Alternatively, it may simply be that our attack methods are suboptimal for deeper encoders.

Figures~\ref{fig:bound_vs_attack_mse}(c--d) show the bound from Corollary~\ref{thm:m_bound2}, with no randomness in the encoder.
Again, the bound always holds, albeit often loose.
Here, weights with skewness controlled by $\alpha$ (green markers) are not shown, because this bound is controlled only by the null space of $\E\big[G\big]$, and the skewness-controlled linear encoders are still full rank. 
%
In this sense, the zero-noise bound (Corollary~\ref{thm:m_bound2}) is inherently looser than the bound for noisy encoders (Corollary~\ref{thm:m_bound1})---Corollary~\ref{thm:m_bound2}, unlike Corollary~\ref{thm:m_bound1}, cannot capture the information loss due to the skewed spectral geometry, unless it introduces a null space.
%
%
%
%

\textbf{LPIPS bound.}
Figures~\ref{fig:bound_vs_attack_lpips}(a--b) show the LPIPS bound from our Theorem~\ref{thm:psi_bound1} versus the actual LPIPS achieved by running the attack. Figures~\ref{fig:bound_vs_attack_lpips}(c--d) show the LPIPS bound when no noise is added to the encoder output (Corollary~\ref{thm:psi_bound2}).
The bounds in Figure~\ref{fig:bound_vs_attack_lpips} always hold, but are generally looser than the MSE bound (Figure~\ref{fig:bound_vs_attack_mse}); this may again be improved by selecting a better set of projectors in Theorem~\ref{thm:psi_bound1} or with a better attack.
%

%
%


%



\begin{figure}[t]
    \centering
    \includegraphics[width=0.45\textwidth]{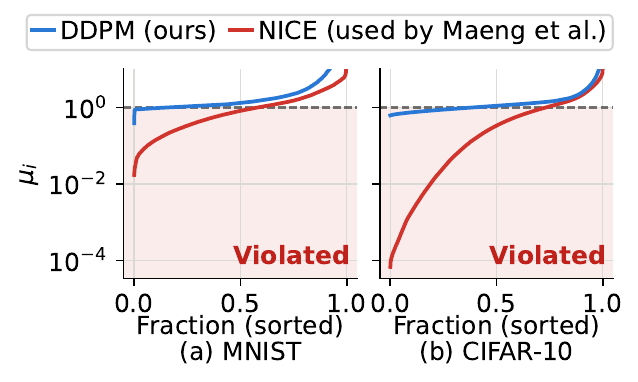}
    \caption{$S$ quality check: all the eigenvalues $\mu_i$ of $\Sigma^{1/2} S \Sigma^{1/2}$ must be equal to or larger than 1.}
    \label{fig:S_eval}
\end{figure}

\subsection{Sensitivity Study}
\label{sec:eval_sensitivity}

In this section, we evaluate the heuristics from Section~\ref{sec:practical_considerations}.

\begin{figure*}[t]
    \centering
    \includegraphics[width=0.9\textwidth]{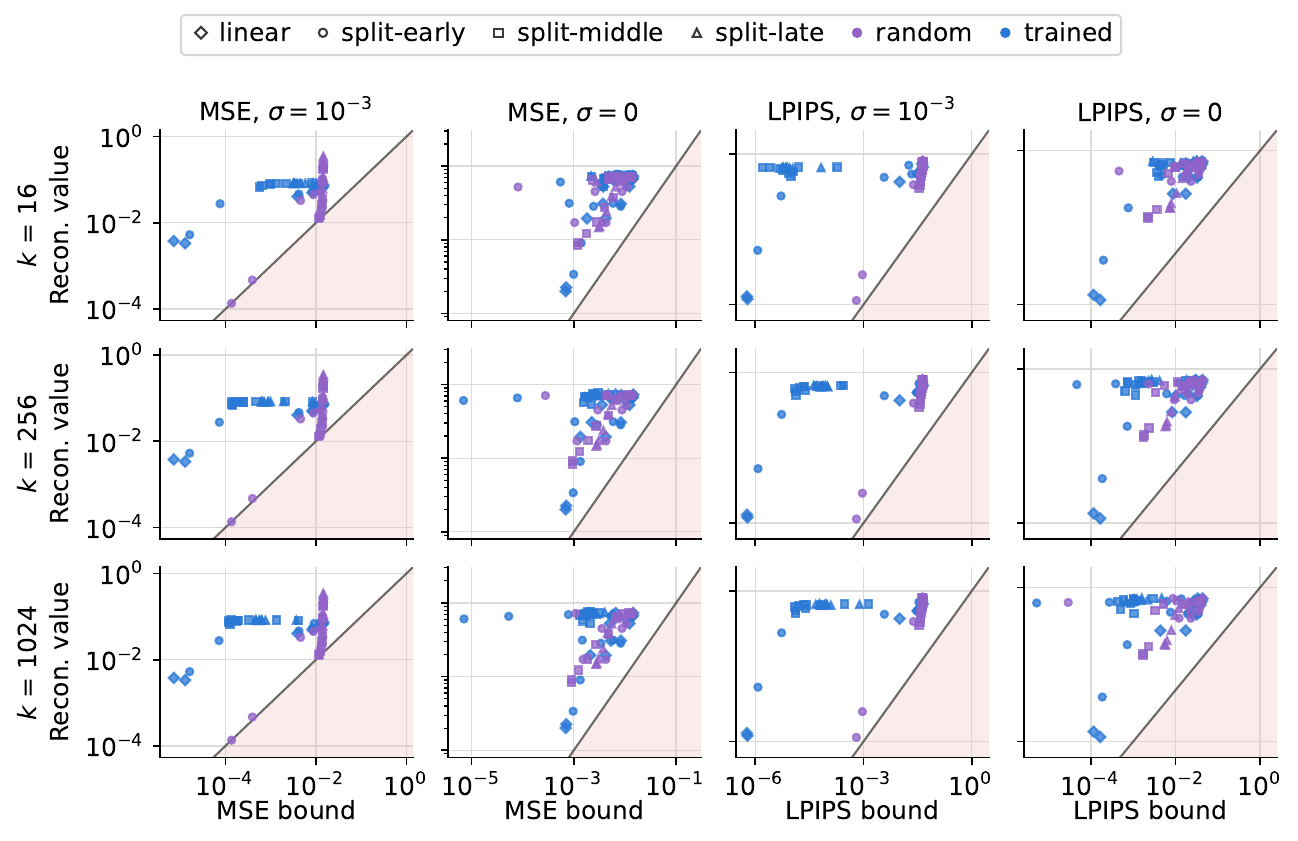}
    \caption{Bound vs. attack MSE/LPIPS with while varying the number of data $k$ used to estimate $\E\big[G\big]$. Result from the CIFAR-10 dataset; MNIST dataset shows similar trends and is omitted for space.}
    \label{fig:k_grid_cifar}
\end{figure*}

\subsubsection{Using Different Score Models}
\label{sec:eval_S_estimation}

We introduced a new practical method for calculating $S=\E_\pi\!\big[s_\pi(x) s_\pi(x)^\top\big]$ using a well-trained DDPM (Section~\ref{sec:heuristics_s_estimation}).
Previously, Maeng et al.~\cite{maeng2023bounding} used a technique called sliced score matching~\cite{sliced_score_matching} to train a NICE model~\cite{nice} to calculate the same matrix. Here, we compare the two approaches, showing that our new method is more reliable.
However, we do not claim that our method is fundamentally superior; rather, we believe the better estimation results are because open-source DDPMs have been well-trained over time by the community, and have therefore internalized better score models than one could obtain by training a score model from scratch.

We measure the quality of $S$ as follows. Since $S$ is the Fisher information of $x \sim \pi$, the matrix form of the Cram\'er-Rao inequality~\cite{dembo1991information} gives $S \succeq \Sigma^{-1}$, where $\Sigma$ is the covariance of $x \sim \pi$ and $\succeq$ indicates the Loewner order. Multiplying both sides by $\Sigma^{1/2}$ gives $\Sigma^{1/2} S \Sigma^{1/2} \succeq I_d$. Since $\Sigma^{1/2} S \Sigma^{1/2}$ is symmetric positive semi-definite, $\Sigma^{1/2} S \Sigma^{1/2} \succeq I_d$ holds if and only if all of its eigenvalues $\mu_i$ satisfy $\mu_i \ge 1$ for all $i$. This is exactly what we check.

We calculate $\Sigma$ from the training images and run this test on $S$ obtained from both our method and the method of prior work~\cite{maeng2023bounding}.
Figure~\ref{fig:S_eval} shows the results. While the eigenvalues ($\mu_i$) from our approach (blue line) also sometimes fall below 1, the violation is much rarer and milder than that of prior approach (red line).
This shows that our DDPM-based method more closely estimates the true $S$.

\subsubsection{Effect of Sample Count $k$ in Estimating $\E\big[G\big]$}
\label{sec:eval_sensitivity_eg_k}

The most accurate way to estimate $\E\big[G\big]$ is to calculate $G = J_{\mathrm{Enc_D}}(x)^\top J_{\mathrm{Enc_D}}(x)$ for each $x$ and average across all training samples, but this requires heavy compute, especially for deep encoders.
Figure~\ref{fig:k_grid_cifar} shows, for the CIFAR-10 dataset, how the four bounds (Theorem~\ref{thm:psi_bound1} and Corollaries~\ref{thm:m_bound1}, \ref{thm:m_bound2}, and \ref{thm:psi_bound2}) change if we use only $k$ samples randomly selected from the training set instead.
The figure shows that the bound (\emph{i.e.}, the distribution of the markers) does not change significantly beyond $k=256$; thus, we use $k=256$ throughout our evaluation.

\begin{figure*}[t]
    \centering
    \includegraphics[width=0.9\textwidth]{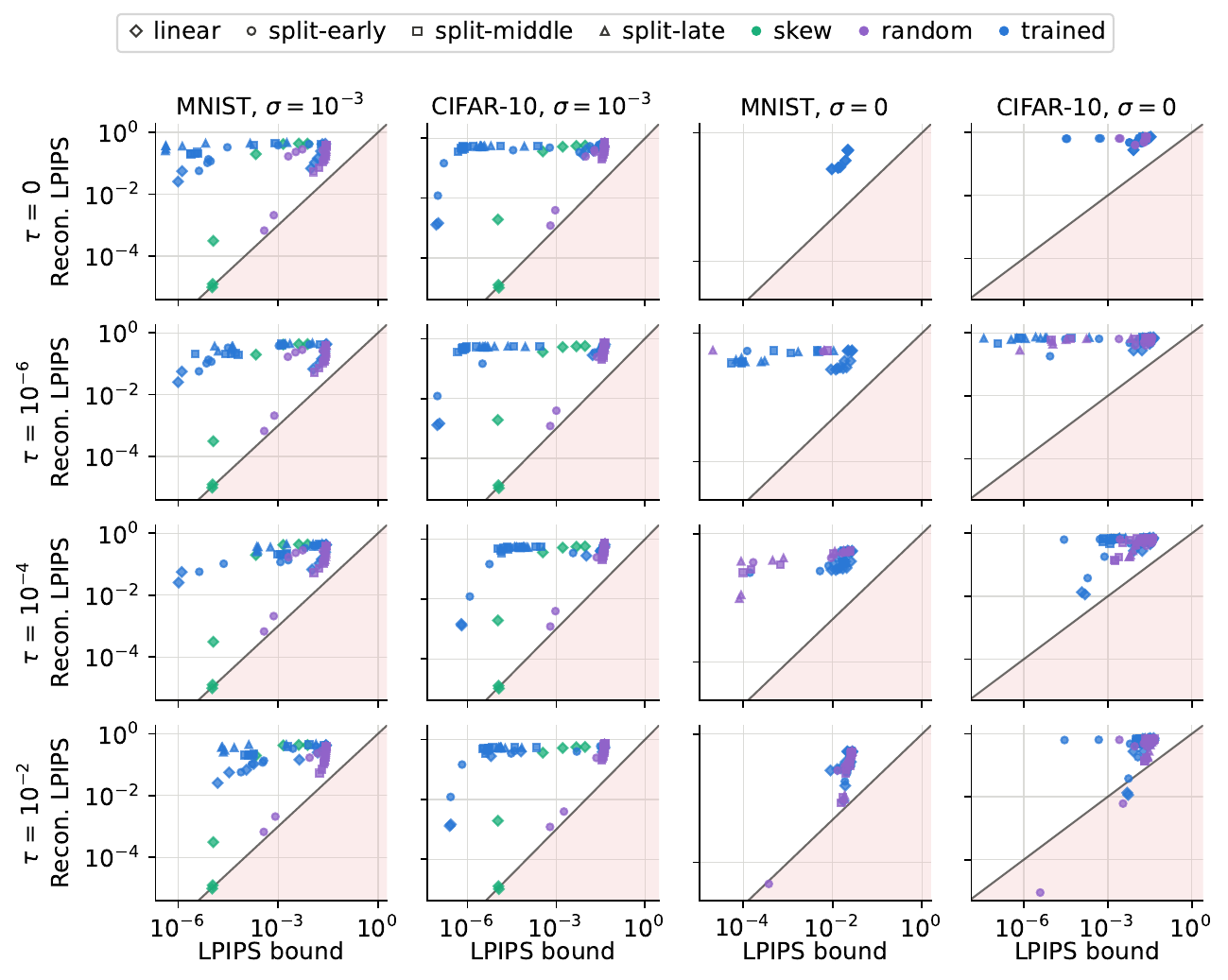}
    \caption{LPIPS bounds with and without noise, under various threshold $\tau$ that splits between the range space and the null space (Section~\ref{sec:eval_sensitivity_eg_tau}). Empirically, $\tau=10^{-4}$ seems to be a reasonable choice, where the bounds appear the tightest and never violated.}
    \label{fig:bound_vs_attack_lpips_tau_sweep}
\end{figure*}

\subsubsection{Effect of Varying the Null Space Threshold $\tau$}
\label{sec:eval_sensitivity_eg_tau}

The bounds from Theorem~\ref{thm:psi_bound1} and Corollaries~\ref{thm:m_bound2} and \ref{thm:psi_bound2} require splitting the domain into the range space and null space of $\E\big[G\big]$.
For Theorem~\ref{thm:psi_bound1}, this split does not need to be accurate---any orthogonal split leads to a valid bound (although its tightness will vary), and using the range and null spaces was simply our choice for simplicity (Section~\ref{sec:psi_bound1}).
However, the bounds for encoders with no randomness (Corollaries~\ref{thm:m_bound2} and \ref{thm:psi_bound2}) require the split to be accurate, since these bounds rely critically on the null space. If part of the range space is incorrectly classified as null, the bound can be overestimated (the reverse is fine, albeit will lead to a looser bound).

Thus, the safest approach is to consider only directions whose eigenvalues are exactly 0 as the null space. However, this may make the bound loose, because our estimate of $\E\big[G\big]$ cannot be exact for real-world data---since it relies on a sampled training dataset---and some null directions' eigenvalues may be estimated as small but nonzero due to estimation noise.
Hence, we treat eigenvalues smaller than $\tau \cdot \lambda_{\mathrm{max}}$ as zero, where $\lambda_{\mathrm{max}}$ is the largest eigenvalue (Section~\ref{sec:eval_sensitivity_eg_tau}). 

Figure~\ref{fig:bound_vs_attack_lpips_tau_sweep} plots the LPIPS bounds for various datasets, with and without noise, to evaluate the effect of choosing different $\tau$. Again, the bounds with noise (Theorem~\ref{thm:psi_bound1}; the first two columns) are always correct, and $\tau$ only controls the tightness.
For these bounds in the first two columns, $\tau=10^{-4}$ and $\tau=10^{-2}$ appear to be the tightest, but the differences across all four rows are not very substantial---\emph{i.e.}, the choice of $\tau$ does not significantly influence the bounds, unless it is too large.

The bounds for zero-noise encoders (Corollaries~\ref{thm:m_bound2} and \ref{thm:psi_bound2}) can become incorrect and be violated if we choose a $\tau$ that is too large.
The figure shows that these bounds are influenced more strongly by the choice of $\tau$, and are indeed violated at $\tau=10^{-2}$. On the other hand, $\tau=0$ is always valid, but appears very loose.
Across the four columns, $\tau=10^{-4}$ appears to be a reasonable choice, giving the tightest bounds that remain correct.
Thus, we use $\tau=10^{-4}$ throughout our evaluation.



%
%
%


\subsection{Visualization of the Reconstruction and Using RTC as a Metric}
\label{sec:eval_rtc}

\begin{figure}[t]
    \centering
    \begin{subfigure}{0.49\textwidth}
        \centering
        \includegraphics[width=\textwidth]{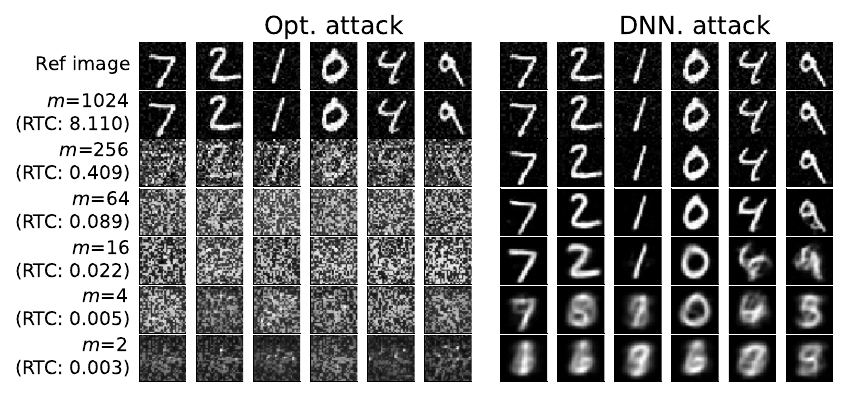}
    \end{subfigure}
    \begin{subfigure}{0.49\textwidth}
        \centering
        \includegraphics[width=\textwidth]{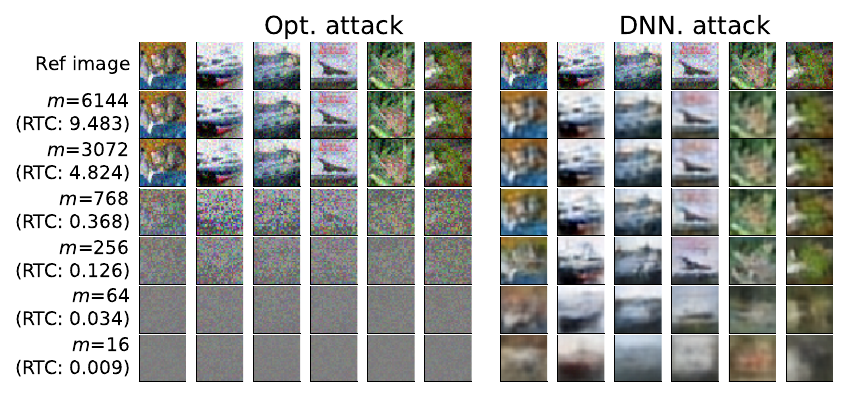}
    \end{subfigure}
    \caption{
    Visualization of the reconstructed images from Figures~\ref{fig:bound_vs_m}(b--c) ($\sigma=10^{-3}$), along with each encoder's RTC.
    }
    \label{fig:recon1}
\end{figure}

\begin{figure}[t]
    \centering
    \begin{subfigure}{0.49\textwidth}
        \centering
        \includegraphics[width=\textwidth]{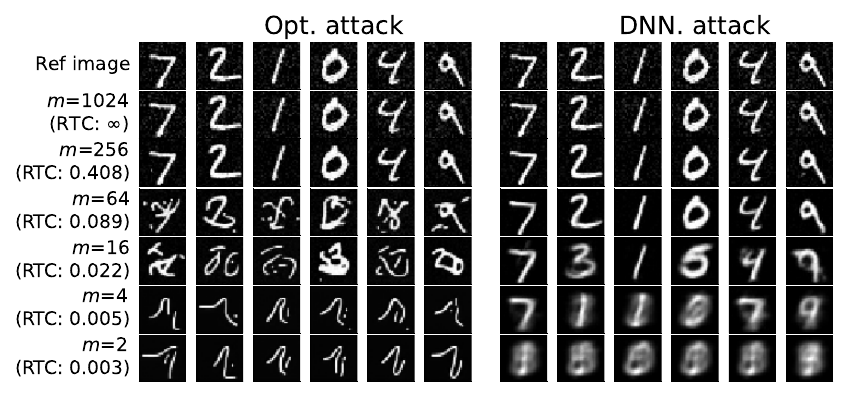}
    \end{subfigure}
    \begin{subfigure}{0.49\textwidth}
        \centering
        \includegraphics[width=\textwidth]{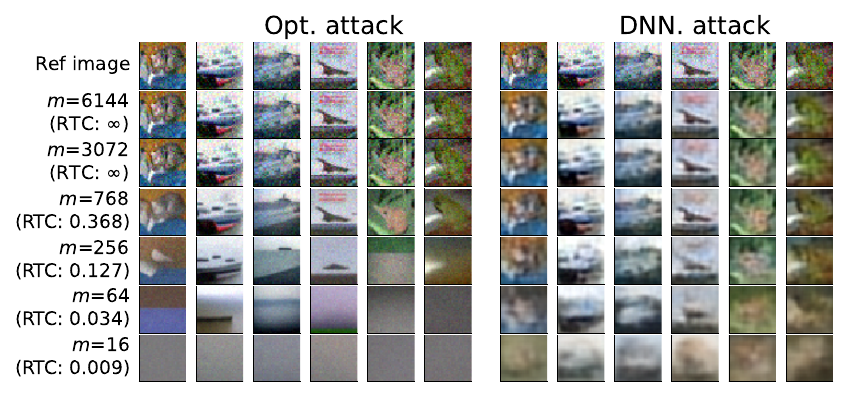}
    \end{subfigure}
    \caption{
    Visualization of the reconstructed images from Figures~\ref{fig:bound_vs_m}(e--f) ($\sigma=0$), along with each encoder's RTC.}
    \label{fig:recon2}
\end{figure}

Figures~\ref{fig:recon1} and \ref{fig:recon2} visualize some reconstructed images from Figures~\ref{fig:bound_vs_m}(b--c) (when the encoder has small randomness) and Figures~\ref{fig:bound_vs_m}(e--f) (when the encoder has no randomness).
We also show the encoder's ratio-to-ceiling (RTC) metric values, an empirical metric we proposed in Section~\ref{sec:ratio_heuristic}.

Smaller embedding dimension $m$---which leads to both increased MSE and LPIPS bounds (Figures~\ref{fig:bound_vs_m} and \ref{fig:bound_psi_vs_m})---also leads to perceptually worse reconstruction.
More interestingly, reconstruction hardness is well captured by the RTC metric: smaller RTC correspond to visually worse reconstruction, and reconstructions with RTC $< 0.01$ fail to recover much meaningful information. These results highlight the potential of RTC as a proxy for invertibility.

The evaluation also shows that MNIST requires a smaller RTC to be strongly protected: even at RTC=0.005, some columns are reasonably reconstructed. This is because the MNIST dataset lies on a much lower-dimensional manifold, which makes reconstruction possible with only very limited  information (discussed further in Section~\ref{sec:why_mnist}).
This implies that the RTC threshold considered safe depends on the dataset and must be empirically determined by the community.
This situation parallels popular privacy definitions like differential privacy (DP)---although DP offers stronger worst-case guarantees, what DP parameters (\emph{e.g.}, $\epsilon$ and $\delta$) are considered safe in practice is still being determined empirically~\cite{jayaraman2019evaluating}, and there is no single number that is universally safe.
%

\subsubsection{Why is MNIST Hard to Perfectly Protect?}
\label{sec:why_mnist}

Figures~\ref{fig:recon1} and \ref{fig:recon2} show that MNIST is sometimes reasonably reconstructed even at a very low RTC (\emph{e.g.}, 0.005). For example, digits 7 and 0 in Figure~\ref{fig:recon1}, and digits 7 and 1 in Figure~\ref{fig:recon2}, are meaningfully reconstructed even at RTC$=0.005$.
This is in contrast to CIFAR-10, whose reconstructed images are much less informative even at a much higher RTC of 0.009.

Our additional study revealed why---MNIST can be reconstructed to a reasonable level much more easily, even with very little information.
We ran a simple experiment to demonstrate this.
We grouped the images in the MNIST training set by their labels and averaged all images sharing the same label, producing a class-mean image for each label.
These class-mean images are what an attacker could ``reconstruct'' by knowing only the label of each image---a mere 3.3 bits ($\log_2(10)$) of information that must survive through the encoder anyway for meaningful downstream classification.

\begin{figure}[t]
    \centering
    \includegraphics[width=0.45\textwidth]{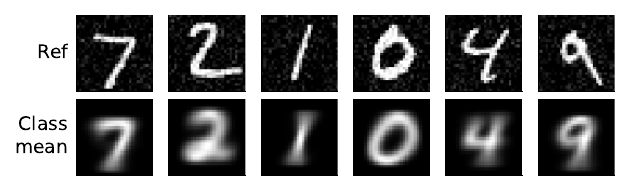}
    \caption{Same reference images from Figures~\ref{fig:recon1} and \ref{fig:recon2}, compared with their class-mean images.
    }
    \label{fig:label_only_mnist}
\end{figure}

Figure~\ref{fig:label_only_mnist} shows the reference images from Figures~\ref{fig:recon1} and \ref{fig:recon2}, along with their respective class-mean images.
The results show that the class-mean image is a surprisingly close perceptual reconstruction of the original image.
This explains why instance encoding cannot work well for the MNIST dataset. The minimal 3.3 bits of label information that the encoder must preserve are sufficient to reconstruct the image, especially for digits like 7, 1, and 0, which were also reasonably well reconstructed in Figures~\ref{fig:recon1} and \ref{fig:recon2}.
This was not the case for CIFAR-10: for example, individual cat images vary enough that the class-mean cat image does not resemble a high-quality reconstruction.
In other words, it is impossible to fully prevent reconstruction (beyond the level shown in Figure~\ref{fig:label_only_mnist}) while still achieving high downstream classification accuracy. If any paper claims such a result, it likely indicates that the attack used is not strong enough.

\section{Conclusion}

We developed a set of bounds that lower-bound the adversary's ability to reconstruct the original data from the embedding, in the setting of instance encoding.
We show that our bounds are tighter than that from prior work~\cite{maeng2023bounding} in many cases, and are useful in cases where the prior bounds do not apply (\emph{e.g.}, when the encoder has no randomness, or when a norm-based metric other than MSE is considered). We also introduce a new, practical way to estimate the data prior more accurately, using a pre-trained DDPM. Our bounds can serve as an indicative metric for an encoder's invertibility.

\section*{Acknowledgments}
We thank Chuan Guo for insightful discussions and pointers on developing the LPIPS bound.
This work was partly supported by the U.S. National Science Foundation under award No. CNS-2349610 and CCF-2529883. Any opinions, findings, and conclusions or recommendations expressed in this material are those of the author(s) and do not necessarily reflect the views of the National Science Foundation. 
The authors used AI for code, text, and figures, but all outputs were manually reviewed by the authors to ensure correctness.

\bibliographystyle{plainurl}
\bibliography{iclr2027_conference}
\appendix
\section{Appendix}
\label{sec:appendix}

\subsection{Evaluation Details}

\subsubsection{Encoder Architecture}
\label{app:arch}

For DNN encoders, we used the following backbone models. We split the backbone models at different split points and added a compression block, as explained in Section~\ref{sec:eval_setup}.
Backbone CNN model for MNIST consist of the following three blocks: 
\begin{itemize}[noitemsep, leftmargin=*, topsep=0pt]
    \item Block 0: Conv2d(out\_dim=32, in\_dim=1, k=5, stride=2, pad=2) + GELU
    \item Block 1: Conv2d(out\_dim=64, in\_dim=32, k=3, stride=2, pad=1) + GELU + AvgPool2d(2)
    \item Block 2: Conv2d(out\_dim=128, in\_dim=64, k=3, stride=1, pad=1) + GELU
    \item Linear(out\_dim=10, in\_dim=128$\times$7$\times$7)
\end{itemize}
Backbone model for CIFAR-10 used the ResNet-18 model from~\cite{resnet18_cifar10}, with MaxPool2d replaced with AvgPool2d and ReLU replaced with GELU.

The compression block controls the output of the encoder $m$ by first changing the channel dimension from $C_{in}$ to $C_{out}$. If this is insufficient (still too large even with $C_{out}=1$) to meet $m$, it reduces the $H$ and $W$ dimensions of the embedding through an additional AdaptiveAvgPool2d layer.
Concretely, to achieve $m=C_{out}\times s$,
\begin{itemize}[noitemsep, leftmargin=*, topsep=0pt]
    \item Compress: Conv2d(out\_dim=$C_{out}$, in\_dim=$C_{in}$, k=3) + GELU + AdaptiveAvgPool2d(s)
    \item Decompress: Upsample(H, W) + Conv2d(out\_dim=$C_{in}$, in\_dim=$C_{out}$, k=3) + GELU
\end{itemize}

\subsubsection{Attack Details}
\label{app:attack_params}

For the Gaussian inputs, each markers are averaged from 64 inputs. For $\sigma > 0$, each input was encoded 8 times and tested. For $\sigma = 0$, each input was encoded and tested once, as everything is fully deterministic.
For MNST and CIFAR-10, we tested 16 inputs, with each encoded and tested once, and averaged for each marker.

For Opt. attack, we use an Adam optimizer. For $\sigma > 0$ case, we set $\beta = 2\sigma^2$ (MAP estimator). For $\sigma=0$, we sweep $\beta \in \{0, 2e^{-6}, 2e^{-4}, 2e^{-3}, 2e^{-2}, 2e^{-1}\}$. We initialize $\hat{x}$ with every pixel being 0.5. We run the optimizer for 500 steps with learning rate 0.05, and CosineAnnealingLR scheduler with $T_{\mathrm{max}}=500$ and $\eta_{\mathrm{min}}=0$.
After each step, the image is clamped to $\hat{x} \in [-0.5, 1.5]$ to refrain from diverging too far from [0, 1].

For DNN attack, we use the following architecture from~\cite{ressfl}, modified to cover arbitrary $m$:
\begin{itemize}[noitemsep, leftmargin=*, topsep=0pt]
    \item First layer: Linear(out\_dim=1024, in\_dim=$m$) + LeakyReLU(0.2) + Linear(out\_dim=128$\times h^2$, in\_dim=1024) + Reshape to (128, $h$, $h$)
    \item Body: 3$\times$[Conv2d(out\_dim=128, in\_dim=128, k=3, stride=1, pad=1) + LeakyReLU(0.2)]
    \item Final layer: ConvTranspose2d(out\_dim=128, in\_dim=128, k=3, stride=2, pad=1, out\_pad=1) + LeakyReLU(0.2) + ConvTranspose2d(out\_dim=C, in\_dim=128, k=3, stride=2, pad=1, out\_pad=1)
\end{itemize}
where $C$ is the channel dimension of the image dataset.
We used $h=7$ for MNIST and $h=9$ for CIFAR-10. For training, we used the entire training set with the following hyperparameters: batch size 256, 40 epochs, Adam with OneCycleLR (max\_lr=2e-3, pct\_start=0.3), plain MSELoss, and clip\_grad\_norm(0.1).
We scaled down the embedding as the scale varies with $m$, by the embeddings' root-mean-square (RMS) value from a sampled 512 embeddings from the training set.


\subsection{Van Trees Inequality (in the form from Bouchard et al.~\cite{bouchard2024intrinsic})}
\label{app:vantrees_bouchard2024}

Below, we restate the matrix van Trees inequality from Bouchard et al.~\cite{bouchard2024intrinsic} (with notational changes) used to derive our Corollary~\ref{thm:m_bound1}. Note that this form is not original to Bouchard et al.~\cite{bouchard2024intrinsic} and similar forms were presented in other work as well, but we use the notation from Bouchard et al.~\cite{bouchard2024intrinsic} as it most closely resemble our need.

\begin{lemma}[Matrix van Trees inequality~\cite{bouchard2024intrinsic}]
\label{thm:vantrees1}
Let $\hat{\theta}$ be an estimator of $\theta$, and $y$ be a noisy observation. 
Let
\[
\mathcal{I}_y(\theta) = \E_{y|\theta}\big[\nabla_\theta \log p(y|\theta) \nabla_\theta \log p(y|\theta)^\top\big],
\]
and $S = \E_{\theta}\big[\nabla_\theta \log \pi(\theta) \nabla_\theta \log \pi(\theta)^\top\big]$, where $\pi(\theta)$ is the probability density function of $\theta$. 
If $p(y, \theta)$ is absolutely continuous with respect to $\theta$ a.e. $y$, if $\theta p(y, \theta) \to 0$ as $\|\theta\| \to \infty$, and if $\E_\theta\big[\mathcal{I}_y(\theta)\big] + S$ exists and is non-singular, 
For a $d$-dimensional $\theta$, the following holds for any $\hat{\theta}$:
\begin{equation}
    \mathbb{E}\big[||\hat{\theta} - \theta||_2^2/d\big] \ge
\frac{1}{d}\Tr\Big(\big(\E\big[\mathcal{I}_y(\theta)\big] + S\big)^{-1}\Big).
\label{eq:bound1}
\end{equation}
\end{lemma}

\subsection{Generalized Multivariate van Trees Inequality (in the form from Gill \& Levit~\cite{gill1995applications})}
\label{app:vantrees_gill1995}

Below, we restate the generalized multivariate van Trees Inequality in the form presented in Gill \& Levit~\cite{gill1995applications}, which we use to derive Theorem~\ref{thm:psi_bound1} and Corollary~\ref{thm:psi_bound2}. Specifically, below is their Theorem 1 with notational changes and simplifications (we set $B=I_k$ and $n=1$) to fit our purpose.

\begin{lemma}[Generalized van Trees with $\Psi$~\cite{gill1995applications}]\label{lem:gill-levit}
Let $\hat{\theta}$ be an estimator of $\theta$, let $y$ be a noisy
observation, and let $\Psi : \R^{d} \to \R^{k}$ be differentiable with
$J_{\Psi} = \partial \Psi / \partial \theta \in \R^{k \times d}$.
Suppose $p(y \mid \theta)$ and $\pi(\theta)$ are absolutely continuous
in $\theta$, and that $\mathcal{I}_{y}(\theta)$ exists with
$\mathrm{diag}(\mathcal{I}_{y}(\theta))^{1/2}$ locally integrable in
$\theta$.  Let $s_{\pi}(\theta) = \nabla_{\theta} \log \pi(\theta)$ be
the score function.  Let $C : \R^{d} \to \R^{k \times d}$ be a
differentiable matrix weight field with rows $c_{i}(\theta)^{\top}$,
with all moments below finite, and suppose
$C(\theta)\, p(y, \theta) \to 0$ as $\|\theta\| \to \infty$.  Then,
\begin{align*}
\E\big[\| &\Psi(\hat{\theta}) - \Psi(\theta) \|^{2} \big] \\
&\ge \frac{\Big( \E\big[ \Tr\big( J_{\Psi}(\theta)\,
    C(\theta)^{\top} \big) \big] \Big)^{2}}
{\E\big[ \Tr\big( C(\theta)\, \mathcal{I}_{y}(\theta)\,
    C(\theta)^{\top} \big) \big]
 + \E\big[ \| \divg C + C\, s_{\pi} \|^{2} \big]},
\end{align*}
where $(\divg C)_{i} = \nabla \!\cdot\! c_{i}$.
\end{lemma}


\subsection{Proofs}
\label{sec:proofs}

\subsection{Proofs of the Bounds}
\printProofs

\subsection{Proofs of the MMSE Attack}

Here, we prove that Equations~\ref{eq:gaussian_mmse} and \ref{eq:gaussian_mmse_zero_noise} are the MMSE estimators when reconstructing Gaussian inputs $x \sim \mathcal{N}(\mu, \eta^2I_d)$ from a linear encoder output $e = Wx + \mathcal{N}(0, \sigma^2I_d)$.

\textbf{$\sigma > 0$ case.}
First, recall that the MMSE estimator is the posterior mean $\hat{x} = \mathbb{E}[x\mid e]$.
For $\sigma>0$,
\[
e\mid x
\sim
\mathcal{N}(Wx,\sigma^2I_m),
\]
so the likelihood is
\[
p(e\mid x)
\propto
\exp\left(
-\frac{1}{2\sigma^2}\|e-Wx\|_2^2
\right).
\]
The prior is
\[
\pi(x)
\propto
\exp\left(
-\frac{1}{2\eta^2}
\|x-\mu\mathbf{1}\|_2^2
\right).
\]
By Bayes' rule,
\[
p(x\mid e)
\propto
p(e\mid x)\pi(x),
\]
and hence
\[
p(x\mid e)
\propto
\exp\left[
-\frac{1}{2\sigma^2}\|e-Wx\|_2^2
-\frac{1}{2\eta^2}\|x-\mu\mathbf{1}\|_2^2
\right].
\]
Expand the two quadratic terms:
\[
\|e-Wx\|_2^2
=
e^\top e
-2x^\top W^\top e
+x^\top W^\top Wx,
\]
and
\[
\|x-\mu\mathbf{1}\|_2^2
=
x^\top x
-2\mu\mathbf{1}^\top x
+\text{constant}.
\]
Ignoring terms that do not depend on $x$,
\[
\begin{aligned}
p&(x\mid e)
\propto\\
&\exp\left[
-\frac{1}{2}
\left(
x^\top
\left(
\frac{W^\top W}{\sigma^2}
+
\frac{I_d}{\eta^2}
\right)
x
-
2x^\top
\left(
\frac{W^\top e}{\sigma^2}
+
\frac{\mu\1}{\eta^2}
\right)
\right)
\right].
\end{aligned}
\]
Define
\[
R
=
\frac{W^\top W}{\sigma^2}
+
\frac{I_d}{\eta^2}
\]
and
\[
b
=
\frac{W^\top e}{\sigma^2}
+
\frac{\mu\mathbf{1}}{\eta^2}.
\]
Then
\[
p(x\mid e)
\propto
\exp\left[
-\frac{1}{2}
\left(
x^\top Rx-2x^\top b
\right)
\right].
\]
As $R$ is symmetric,
\[
x^\top Rx-2x^\top b
=
(x-R^{-1}b)^\top R(x-R^{-1}b)
+\text{constant},
\]
so
\[
p(x\mid e)
\propto
\exp\left[
-\frac{1}{2}
\left((x-R^{-1}b)^\top R(x-R^{-1}b)
\right)
\right].
\]
This is a Gaussian with mean $R^{-1}b$ and covariance $R^{-1}$:
\[
x\mid e
\sim
\mathcal{N}(R^{-1}b,R^{-1}).
\]
Thus, the posterior mean (MMSE) is
\[
\hat{x}
=
R^{-1}
\left(
\frac{W^\top e}{\sigma^2}
+
\frac{\mu\mathbf{1}}{\eta^2}
\right).
\]
Using
\[
R\mu\mathbf{1}
=
\frac{W^\top W\mu\mathbf{1}}{\sigma^2}
+
\frac{\mu\mathbf{1}}{\eta^2},
\]
this can equivalently be written as
\[
\hat{x}
=
\mu\mathbf{1}
+
R^{-1}
\frac{W^\top (e-W\mu\mathbf{1})}{\sigma^2},
\]
which is Equation~\ref{eq:gaussian_mmse}.

\textbf{$\sigma = 0$ case.}
When $\sigma=0$, the above equation cannot be used because of the $1/\sigma^2$ term. Instead, note that the observation is noiseless: $e=Wx$.
Define $y=x-\mu\mathbf{1}$.
Then,
\[
y\sim\mathcal{N}(0,\eta^2I_d),
\]
and
\[
e-W\mu\mathbf{1}=Wy = z.
\]
Our goal is to find
\[
\hat{x} = \E\big[x | e\big] = \E\big[y | e\big] + \mu\1 = \E\big[y | z\big] + \mu\1.
\]
Let $P_\mathrm{row}$ and $P_\mathrm{null}$ be the orthogonal projectors to row and null spaces of $W$.
Decompose $y$ into its row-space and null-space components:
\[
y
=
P_\mathrm{row}y
+
P_\mathrm{null}y.
\]
The null-space component is completely invisible to the observation,
since
\[
WP_\mathrm{null}y=0.
\]
Moreover, because $y\sim\mathcal{N}(0,\tau^2I)$, the row-space and
null-space components are independent. Therefore, observing $z$ provides
no information about $P_\mathrm{null}y$, and
\[
\E\big[P_\mathrm{null}y\mid z\big]
=
\E\big[P_\mathrm{null}y\big]
=
0.
\]
For the row-space component, we use
\[
P_\mathrm{row}=W^+W,
\]
where $W^+$ is a Moore-Penrose pseudoinverse of $W$.
Thus,
\begin{align*}
\E\big[y\mid z\big]+\mu\mathbf{1}
&=
\E\big[P_\mathrm{row}y\mid z\big]+\mu\mathbf{1}\\
&=
\E\big[W^+Wy\mid z\big]+\mu\mathbf{1}\\
&=
\E\big[W^+z\mid z\big]+\mu\mathbf{1}\\
&=
W^+z+\mu\mathbf{1}\\
&=
W^+(e-W\mu\mathbf{1})+\mu\mathbf{1}.
\end{align*}
This is Equation~\ref{eq:gaussian_mmse_zero_noise}, thus we conclude our proof.

\end{document}